%% file: RAL2020-Main.tex
\documentclass[letterpaper, 10 pt, conference]{ieeeconf}  % Comment this line out if you need a4paper

\IEEEoverridecommandlockouts                              % This command is only needed if 
\usepackage{graphics} % for pdf, bitmapped graphics files
\usepackage{epsfig} % for postscript graphics files
\usepackage{mathptmx} % assumes new font selection scheme installed
\usepackage{amsmath} % assumes amsmath package installed
\usepackage{amssymb}  % assumes amsmath package installed
\usepackage{graphicx}
\usepackage{bm}
\usepackage{eucal}
\usepackage{todonotes}[inline]  

\usepackage{float}
\usepackage[ruled,vlined]{algorithm2e}
\usepackage{pgfplotstable}
\usepackage{pgfplots}
\usepackage{tikz}
\usetikzlibrary[pgfplots.groupplots]
\usepgfplotslibrary{fillbetween}
\pgfplotsset{compat=1.5}
\usepackage{subcaption}
\usepackage{caption}
\usepackage{xcolor}
\usepackage{enumerate}
\usepackage{makecell}
\usepackage{colortbl}

\usepackage{graphicx}
\usepackage{tabularx,ragged2e,booktabs}
\usepackage{url}
\usepackage{siunitx}
\usepackage{rotating}
\usepackage{amssymb}
\usepackage{tabularx}
\usepackage{pgfplots}
\usepackage{csvsimple}
\usepackage{multirow}
\usepackage{cite}
\usepackage{float}
\usepackage{graphicx}
\usepackage{booktabs}

\DeclareMathOperator*{\argmax}{arg\,max}

\title{\LARGE \bf
Active Visuo-Tactile Point Cloud Registration for Accurate Pose Estimation of Objects in an Unknown Workspace 
}
\author {Prajval Kumar Murali, Michael Gentner, and Mohsen Kaboli% <-this % stops a space
\thanks{P.K.Murali, M. Gentner, M.Kaboli are with the BMW Group, Munich Germany. 
e-mail: name.surname@bmwgroup.com}%
\thanks{P.K. Murali is with the University of Glasgow, Scotland}%
\thanks{M. Gentner is with the Technische Universität München, Germany}
\thanks{M. Kaboli is with the Donders Institute for Brain and Cognition, Radboud University, Netherlands }%
}
\begin{document}

\maketitle
\thispagestyle{empty}
\pagestyle{empty}
% \IEEEpubidadjcol

\begin{abstract}
This paper proposes a novel active visuo-tactile based methodology wherein the accurate estimation of the time-invariant $SE(3)$ pose of objects is considered for autonomous robotic manipulators. The robot equipped with tactile sensors on the gripper is guided by a vision estimate to actively explore and localize the objects in the unknown workspace. The robot is capable of reasoning over multiple potential actions, and execute the action to maximize \textit{information gain} to update the current belief of the object. We formulate the pose estimation process as a linear translation invariant quaternion filter (TIQF) by decoupling the estimation of translation and rotation and formulating the update and measurement model in linear form. We perform pose estimation sequentially on acquired measurements using very sparse point cloud ($\leq 15$ points) as acquiring each measurement using tactile sensing is time consuming. Furthermore, our proposed method is computationally efficient to perform an exhaustive uncertainty-based active touch selection strategy in real-time without the need for trading information gain with execution time.
We evaluated the performance of our approach extensively in simulation and by a robotic system.

\end{abstract}

\input{sections/introduction.tex}
\input{sections/methods.tex}

\input{sections/experiments.tex}

\input{sections/conclusions.tex}

\section*{ACKNOWLEDGMENT}
\input{sections/acknowledgement.tex}

\bibliography{IEEEexample}
\bibliographystyle{IEEEtran}

\end{document}

%% file: sections/introduction.tex
\section{INTRODUCTION}
\label{sec:introduction}
% \begin{itemize}
%     \item Motivation:
%     \begin{itemize}
%         \item Object pose estimation is crucial for robotic grasping. Manipulation in unstructured environments requires both vision and tactile sensing.
%         \item Using vision alone for object pose estimation can fail due to: limitations of the sensor, incorrect calibration, object properties (specular/transparent), occlusions or incorrect point-cloud registration.
%         \item The pose estimation can fail without notice i.e., an faulty sensor can return valid information or pose estimation algorithm returns a solution that corresponds to local-minima. In a real-scenario where there is no information about the ground-truth, there is no way to \textit{verify} or \textit{correct} this pose estimation solution without the use of another modality \cite{navarro2016iso10218}.
        
%     \end{itemize}
% \end{itemize}

%% Motivation
Accurate estimation of object pose (translation and rotation) is crucial for autonomous robots to grasp and manipulate objects in an unstructured environment. Even small inaccuracies in the belief of the object pose can generate incorrect grasp configurations and lead to failures in manipulation tasks~\cite{Qiang-TRO-2020}. 
Strategies based on vision sensors are commonly used for estimating the pose of the object, but there is residual uncertainty in the estimated pose due to incorrect calibration of the sensors, environmental conditions (occlusions, presence of extreme light, and low visibility conditions), and object properties (transparent, specular, reflective). Tactile sensors in combination with robot proprioception provides high fidelity local measurements regarding object pose. However, mapping entire objects using tactile sensors is highly inefficient and time-consuming which necessitates the use of intelligent data gathering strategies and combining vision sensing to drive the tactile sensing~\cite{hsiao2011robust}.

%\begin{figure}[t!]
%    \centering
%    \includegraphics[width = \columnwidth]{figures/experiment_setup.png}
%   \caption{Experimental Setup}
%   \label{fig:expsetup}
%\end{figure}

\begin{figure}[t!]
\centering
   \includegraphics[width = \columnwidth, height = 6cm, trim=1cm 0.1cm 0.5cm 0.1cm, clip=true]{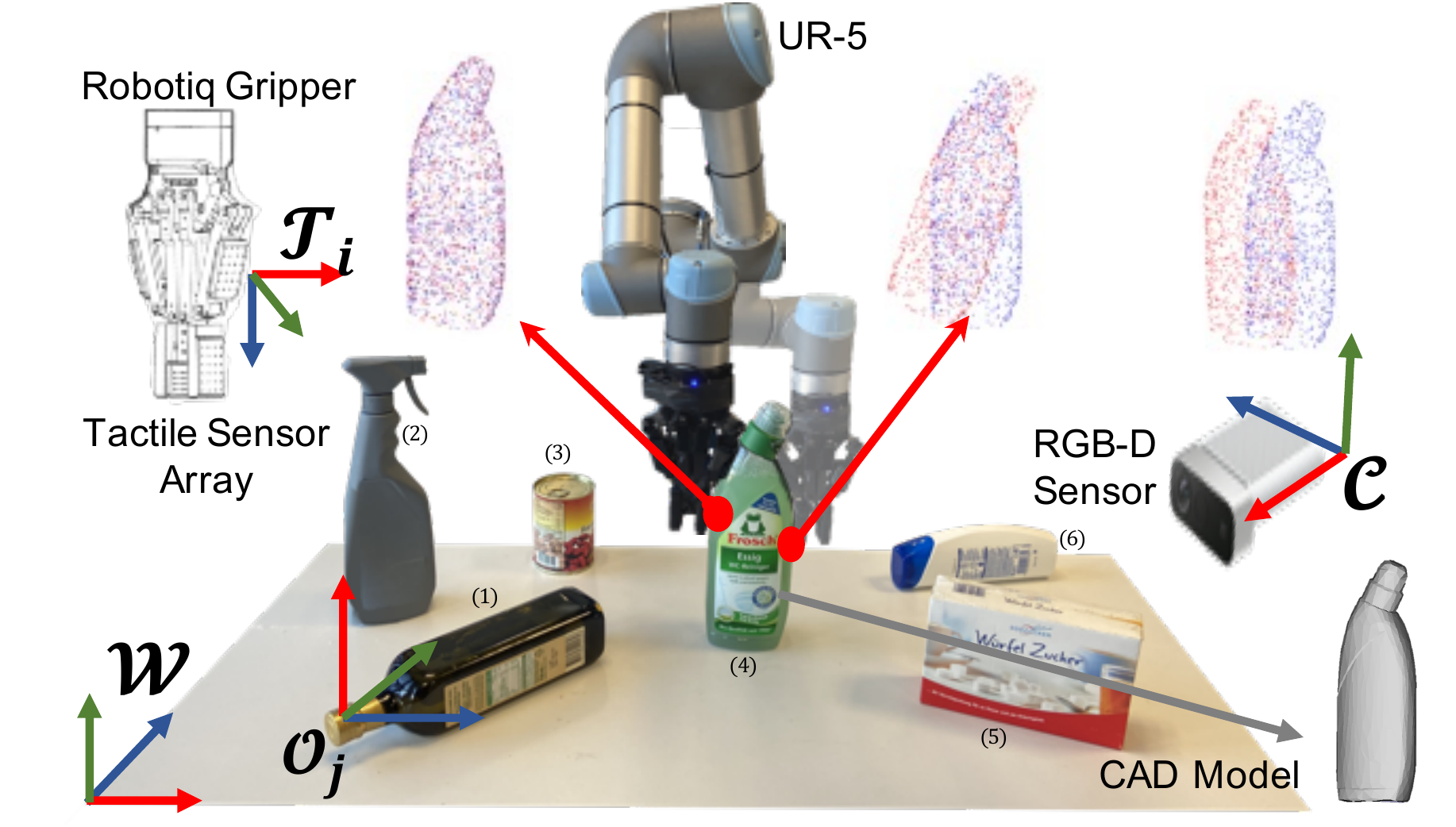}
   \caption{Experimental setup. A Robotiq two-finger adaptive robot
gripper is equipped with 3-axis tactile sensor arrays and mounted on a UR5 robotic arm. In this figure, 6 experimental objects are selected and
placed in the workspace. The experimental objects constitute daily objects as follows: (1) olive oil bottle, (2) spray, (3) can, (4) cleaner, (5) sugar box, (6) shampoo. In experiments, objects were placed in the workspace with various locations and orientations.}
   \label{fig:real_robot}
\end{figure}

%% State-of-art
% The problem of finding the 6 Degree-of-Freedom (DoF) object pose known as object pose estimation or object localisation relying on vision and tactile sensing has been extensively studied in literature \cite{miller1999integration, prats2009vision, hebert2011fusion}. 
Point cloud registration is the process of finding the rigid transformation that aligns two point clouds which is often used for object pose estimation. 
% In the case of pose estimation, the problem deals with aligning one point cloud extracted using external sensors often called the \textit{scene cloud} with the point cloud obtained from the geometric model of the object of interest i.e., \textit{model cloud}.
When correspondences between the two point clouds are known \textit{a priori}, the registration problem can be solved deterministically~\cite{horn1987closed}. However, correspondences are unknown in practical situations and classical approaches involve iteratively finding the best correspondence and the transformation given the best correspondences known as the iterative closest point (ICP) algorithm~\cite{besl1992method}. ICP and its variants~\cite{pomerleau2013comparing} are batch registration methods that are known to have high computation times, and low performance when sparse data is available that arrive sequentially as is the case with tactile measurements~\cite{glozman2001surface}. Hence filter-based approaches are generally preferred for sequential data~\cite{arun2019registration}.
A particle filter based technique, named Scaling Series which localises the object using touch sensing efficiently and reliably was developed in~\cite{petrovskaya2011global}. 
Similarly in~\cite{vezzani2017memory}, the authors proposed the Memory Unscented Particle Filter (MUPF) drawing inspiration from the Unscented Particle Filter (UPF)~\cite{van2001unscented} to localise the object recursively given contact point measurements. 
Alongside contact point measurements based localization, array-based tactile sensors have been used to extract local geometric features of objects using principal component analysis (PCA) in order to localise the object by matching the covariances of the extracted tactile data and object geometry~\cite{bimbo2016hand}. 
Other works have developed specialised tactile descriptors to extract robust tactile information regardless of the nature of tactile sensing and method of exploration~\cite{kaboli2018robust}.
Vision has been used to provide an initial estimate of the object pose that is finely refined by tactile localisation using local or global optimization techniques~\cite{hebert2011fusion}. 
Tactile measurements are inherently sparsely distributed and a probabilistic method was proposed in~\cite{arun2019registration} to perform registration given sparse point cloud, surface normal measurements and the geometric model of the object.
% As noted by~\cite{luo2017robotic} high accuracy and real-time performance is yet to be achieved for localisation of objects using tactile or visuo-tactile sensing approaches. 
While tactile data can be collected in a randomised manner~\cite{arun2019registration} or driven by a human-teleoperator~\cite{vezzani2017memory}, active touch strategies which allows for autonomous data collection and reduction of redundant data collection are required~\cite{kaboli2019tactile}.
Several works have used information gain metric based on the uncertainty of the object’s pose to determine the next best touching action to localise the object~\cite{kaboli2017tactile,hebert2013next,saund2017touch}.
For instance, metrics such as Shannon Entropy have been used to select the next best tactile exploratory action in order to reduce the robot's uncertainity regarding the object properties~\cite{kaboli2018active, feng2018active}.
% Hsiao et al.~\cite{hsiao2010task} implemented a decision-theoretic approach and an approximate Partially Observable Markov Decision Process (POMDP) to select actions for exploration. 
As computation of the next best touch can be computationally expensive, some works have constrained the optimization problem by including the computation and action execution time~\cite{tosi2014action}.

% The authors proposed a deterministic method using iterative most likely oriented point (IMLOP) algorithm~\cite{billings2014iterative} and probabilistic method using the dual quaternion filter (DQF) approach~\cite{srivatsan2016estimating}. 
The key gaps in research are as follows: typical filter-based tactile localisation methods require large number of measurements or predetermined location of touches which may be impractical, time-consuming and requires a human-in-the-loop for manipulation. Furthermore, object localisation using \textit{sparse} measurements has only been reported using random touch selection or manual teleoperated strategies. Current strategies for action touch selection based on expected belief state is computationally intensive due to the use of non-parametric methods such as particle filters and it introduces time delays, while otherwise more information could be obtained through less-optimal but computationally inexpensive random measurements. Moreover, current approaches reason over a \textit{pre-defined} set of actions for the expected information-gain which affects the adaptability of the methods to unstructured scenarios.

%% Contributions
\paragraph*{Contribution}

% {\color{red}{the contribution should be some bullet points. never cite any paper here. most of the text here should be moved to the and of SOA}}
In order to tackle the aforementioned gaps in research, we propose a \textit{novel} framework for active visuo-tactile point cloud registration for accurate pose estimation of objects. Our contributions are as follows:
\newline (I) We propose a translation-invariant quaternion filter (TIQF) for dense-batch vision-based point clouds and sparse-sequential tactile-based point cloud for point cloud registration.
\newline (II) We design an active touch strategy to enable the robot to generate candidate actions and select the optimal action strategically based on information gain. Our active strategy is shown to be computationally efficient to perform an exhaustive uncertainty-based action selection in real-time without the need for trading information gain with execution time. 
The vision pose estimate is corrected by using the tactile modality using the active touch strategy.
\newline (III) We perform extensive experiments in simulation and robotic setup to compare our proposed active strategy against random strategy which is used in state-of-art sparse point cloud registration algorithms.

%% file: sections/methods.tex
\section{METHODS}
\label{sec:methods}

%-Generative model we want to use
%-Show decoupling of rotation and translation (mention that this can be done without even considering dual-quaternions)
%-Show closed form solution for translation
%-We pose the active object localization problem as a bayesian filtering problem
%-Show short derivation of formula with actions
%-Show that this leads to a Kalman filter
%-We decide to use quaternions as smooth representation, since we need a linear model for the kalman filter

%-Thats why we want to use dual-quaternions
%-Write the generative model as dual-quaternion
%-Looking at two pairs of correspondences we can construct translation invariant measurements
%-Therefore we can decouple translation and rotation
%-Write down the generative model for this 
%-Write down the objective
%-This is linear and we have continuously arriving measurements, also we want to reason over how possible actions affect the result
%-Therefore we will incorporate this into a bayesian filter
%-Bayesian approach to object localization (bayesian network)
%-Write down equations, then write down assumptions
%-Then we get to a Kalman filter
%-To incorporate the above generative model we use matrix formulation of this equation we use the concept of pseudo measurements
%-Derive all equations for kalman
%-This leads us to DQF (parametrized by an action)
%-To perform the action selection we will adopt a scheme known from other works
%-How to parametrize the state vector?
%-

\subsection{Problem Formulation}
\begin{figure*}[t!]
    \centering
    \includegraphics[width = 0.9\textwidth, height = 8cm]{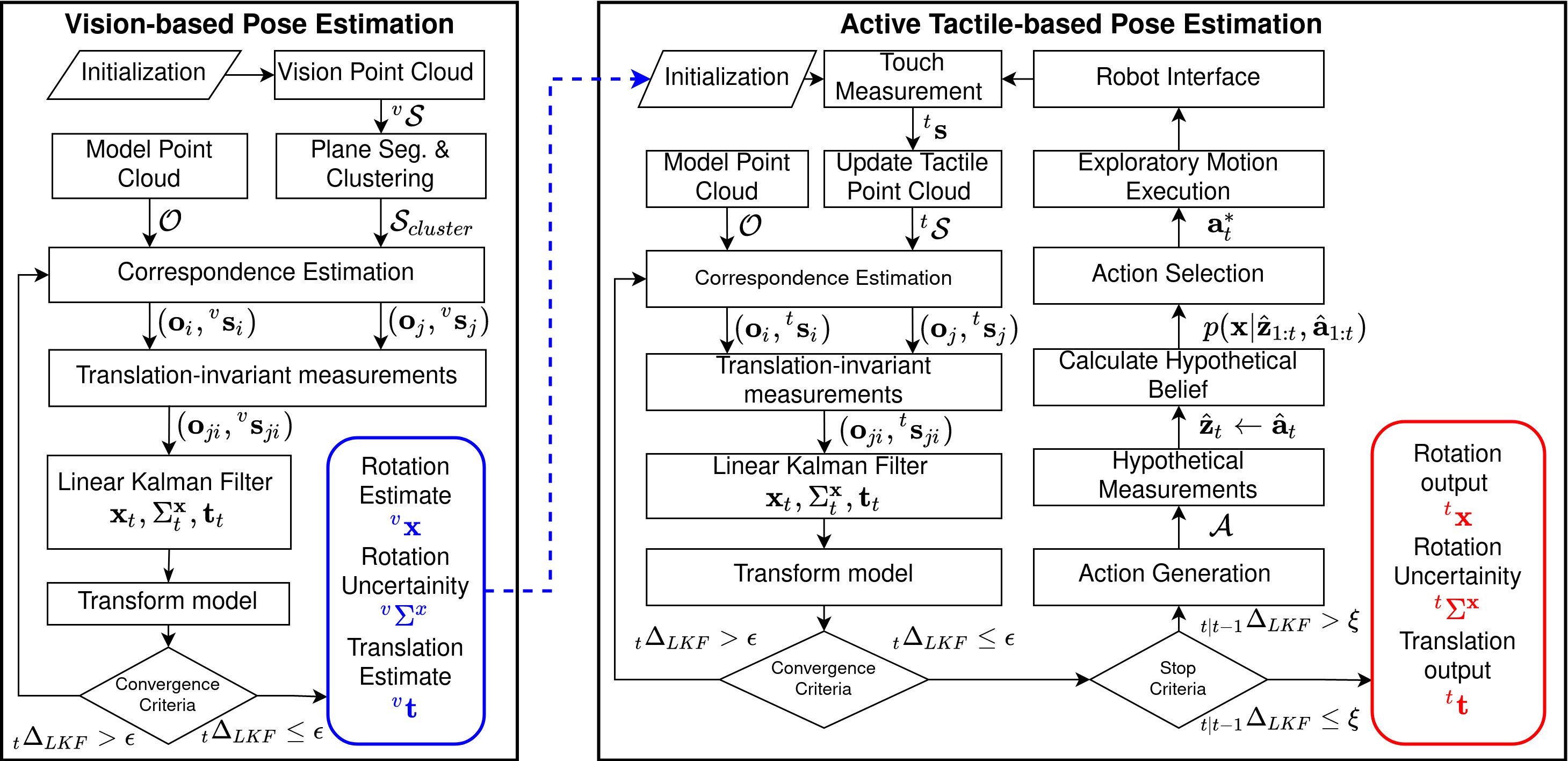}
    \caption{The proposed framework for an active visuo-tactile point cloud registration for the accurate object localization}
    \label{fig:framework}
\end{figure*}
% \todo[inline]{simplify the framework}

We propose an active visuo-tactile based framework shown in Figure~\ref{fig:framework} to perform object pose estimation using point cloud registration. The problem is formally defined as follows: given $N_O$ objects with designated frames $\mathcal{F}_k$ with $k=1,\ldots, N_O$ in the workspace $W_{XYZ}$ of the robot with unknown poses. The workspace $W_{XYZ}$ is defined as a discretised 3D grid bounded by the kinematic reaching constraints of the robot and defined in the world coordinate frame $\mathcal{W}$. We assume the pose of each object is static in time. 
%Furthermore, we also assign for each object a coordinate frame  rigidly attached to the geometric model of the object.
The objective is to find the object pose ${}^{\mathcal{W}} H_{\mathcal{F}_k}$ given sensor measurements ${}^v\mathcal{S}$ from vision sensor and tactile sensor ${}^t\mathcal{S}$. 

\subsection{Proposed Framework}
As described in the framework in Figure~\ref{fig:framework}, vision-based pose estimation is used for providing an initial estimate for the pose estimation by active tactile exploration. The point cloud ${}^v \mathcal{S}$ is captured by the vision sensor and is transformed to the world frame $\mathcal{W}$ by applying the ${}^{\mathcal{W}} H_{\mathcal{C}}$ homogeneous transformation typically known as the hand-eye transformation~\cite{murali2021situ}, where $\mathcal{C}$ is the vision sensor frame. Pre-processing such as plane segmentation and clustering is performed to extract the points corresponding to the object of interest. We propose a translation-invariant quaternion filter (TIQF) which is a probabilistic pose estimation approach which is detailed in Section~\ref{sec:tiqf}. The TIQF filter works on a putative set of correspondences between the model and scene clouds which is found in the correspondence estimation step using the closest point rule~\cite{besl1992method}. The TIQF algorithm upon convergence provides the rotation estimate ${}^v \mathbf{x}$, the corresponding rotation uncertainty ${}^v \Sigma^\mathbf{x}$ and translation estimate ${}^v \mathbf{t}$. 
We use the pose estimate from vision in order to initialise the active tactile-based pose estimation procedure as there can be residual errors in pose estimation from vision-based sensors that can be corrected with high fidelity tactile measurements.
The tactile-based pose estimation is also performed using the TIQF algorithm and it shows the adaptability of the algorithm to handle batch data and sequential data as well as dense and sparse data.
Furthermore, we design an active touch selection strategy as described in Section~\ref{sec:active_selection} in order to intelligently and efficiently extract tactile measurements as performing each tactile measurement is a time-consuming process.

\subsection{Translation-Invariant Quaternion Filter (TIQF)}
\label{sec:tiqf}
To solve the point cloud registration problem for the vision-based pose estimation and the active tactile-based pose estimation in the same manner, we design a linear translation-invariant quaternion filter (TIQF). 
For point clouds from a vision sensor, the TIQF algorithm can be used in a batch manner and during active tactile exploration it can handle sequential point measurements as well. 
The point cloud registration problem given known correspondences can be formalised as follows:
\begin{equation}
     \mathbf{s}_i = \mathbf{R}\mathbf{o}_i + \mathbf{t} \quad i = 1, \dots N \quad ,
     \label{eq:generativemodel}
 \end{equation}
 where $\mathbf{s}_i \in \mathbb{R}^3$ are points belonging to the scene cloud $\mathcal{S}$ drawn from sensor measurements and $\mathbf{o}_i \in \mathbb{R}^3$ are the corresponding points belonging to the model cloud $\mathcal{O}$.
 %generated from the corresponding model mesh assuming no scale variation between the points.
 Rotation and translation are defined as $\mathbf{R} \in SO(3)$ and  $\mathbf{t} \in \mathbb{R}^3$ which are unknown and need to be computed in order to align $\mathbf{o}_i$ with $\mathbf{s}_i$. We decouple the rotation and translation estimation as translation can be trivially computed once rotation is known~\cite{horn1987closed}.
 Given a pair of correspondences $(\mathbf{s}_i, \mathbf{o}_i)$ and $(\mathbf{s}_j, \mathbf{o}_j)$, we define $\mathbf{s}_{ji} = \mathbf{s}_{j} - \mathbf{s}_{i}$ and $\mathbf{o}_{ji} = \mathbf{o}_{j} - \mathbf{o}_{i}$. From Equation~\eqref{eq:generativemodel} we have:
\begin{align}
    \mathbf{s}_j - \mathbf{s}_i &= (\mathbf{R}\mathbf{o}_j + \mathbf{t}) - (\mathbf{R}\mathbf{o}_i + \mathbf{t}) \quad ,\\
    \mathbf{s}_{ji} &= \mathbf{R}\mathbf{o}_{ji}  \quad . 
    \label{eq:trans_invariance}
\end{align}
Equation \eqref{eq:trans_invariance} is independent of $\mathbf{t}$ and once rotation $\hat{\mathbf{R}}$ is estimated, the translation $\hat{\mathbf{t}}$ can be obtained in closed form from Equation \eqref{eq:generativemodel}.
%\begin{equation}
%    \mathbf{t} = \frac{\mathbf{a}_{j}+ \mathbf{a_{i}}}{2} - \mathbf{R} \frac{\mathbf{b_{j}}+ \mathbf{b_{i}}}{2} \quad .
%    \label{eq:trans_output}
%\end{equation}

%\todo[inline]{Work-in-progress from this point onwards}

%\subsubsection{Bayes Filter}

%\begin{figure}[bt]
%%    \centering
%    \includegraphics[width = \columnwidth]{figures/bayesian_network.png}
%    \caption{Bayesian Networks with standard \textit{Kalman Filter} (left) and our model (right)}
%    \label{fig:bayesian_network}
%\end{figure}

We cast the rotation estimation problem into a Bayesian estimation framework.
We define the rotation as the state $\mathbf{x}$ of our filter. Objects are assumed to be fixed, therefore the true state $\mathbf{x}$ is static and does not change over time.
This assumption is realistic considering heavy objects and light contacts during tactile exploration~\cite{petrovskaya2011global}.
During exploration of the workspace by performing actions $a_t$ we obtain tactile measurements $\mathbf{z}_t$. These measurements are then used to update our current belief of the state. Using a Bayesian formulation:
\begin{align}
    p(\mathbf{x} | \mathbf{z}_{1:t}, \mathbf{a}_{1:t}) &= \eta p(\mathbf{x}, \mathbf{z}_{1:t},\mathbf{a}_{1:t}) \\
    &= \eta p(\mathbf{z}_{t} | \mathbf{x}, \mathbf{z}_{1:t-1}, \mathbf{a}_{1:t}) p(\mathbf{x}, \mathbf{z}_{1:t-1}, \mathbf{a}_{1:t}) \quad ,
    \label{eq:bayesian_filter:1}
\end{align}
where $\eta$ is a normalization constant. Since $\mathbf{z}_t$ only depends on the action at timestep $t$ and the state, we can simplify \eqref{eq:bayesian_filter:1} to
\begin{align}
    p(\mathbf{x} | \mathbf{z}_{1:t}, \mathbf{a}_{1:t}) &= \eta p(\mathbf{z}_{t} | \mathbf{x}, \mathbf{a}_{t}) p(\mathbf{x}, \mathbf{z}_{1:t-1}, \mathbf{a}_{1:t}) \\
    &= \eta p(\mathbf{z}_{t} | \mathbf{x}, \mathbf{a}_{t}) p(\mathbf{x}| \mathbf{z}_{1:t-1}, \mathbf{a}_{1:t-1}) 
    \label{eq:bayesian_filter:2}
\end{align}
Note that $p(\mathbf{x}| \mathbf{z}_{1:t-1}, \mathbf{a}_{1:t}) = p(\mathbf{x}| \mathbf{z}_{1:t-1}, \mathbf{a}_{1:t-1})$, since we do not consider the state depending on future actions. 
%\todo[inline]{Dependence on actions is not super clear. Make this simplification more clear}
%With equation \ref{eq:bayesian_filter:2}, we arrive at the formalization for object localization in the static case conditioned on the actions. 
%\begin{equation}
%    bel(\mathbf{x})_t = \eta p(\mathbf{z}_{t} | \mathbf{x}, \mathbf{a}_{t}) bel(\mathbf{x})_{t-1}
%    \label{eq:bayesian_kalman}
%\end{equation}
The dependence of $\mathbf{x}$ on the actions is solemnly stemming from the measurement model $p(\mathbf{z}_{t} | \mathbf{x}, \mathbf{a}_{t})$. 
 
%\subsubsection{Kalman Equations}
% We assume that $\mathbf{x}$ and $\mathbf{z}_t$ are Gaussian distributed.
We choose quaternions as a smooth representation for the state $\mathbf{x}$ and estimate it using a \textit{Kalman Filter}. To leverage the insights from Equation \eqref{eq:trans_invariance}, we formulate a linear measurement model as in~\cite{srivatsan2016estimating}. We can rewrite \eqref{eq:trans_invariance} as: 
\begin{equation}
    \widetilde{\mathbf{s}}_{ji} = \mathbf{x} \odot \widetilde{\mathbf{o}}_{ji} \odot \mathbf{x}^{*} \quad . 
    \label{eq:quat_objective}
\end{equation}
Since $\mathbf{x}$ is a unit quaternion, we use $\sqrt{\mathbf{x}\odot \mathbf{x}^{*}} = ||\mathbf{x}||=1$ to get
\begin{align}
    \widetilde{\mathbf{s}}_{ji}\odot \mathbf{x} &= \mathbf{x} \odot \widetilde{\mathbf{o}}_{ji} \\
    \widetilde{\mathbf{s}}_{ji}\odot \mathbf{x} &- \mathbf{x} \odot \widetilde{\mathbf{o}}_{ji} = 0 \quad .
    \label{eq:quat_objective_2}
\end{align}
%% TODO: explain this step
We can further rewrite~\eqref{eq:quat_objective_2} using the matrix notation of quaternion multiplication as: 
\begin{align}
    \begin{bmatrix}
        0 & -\mathbf{s}{ji}^T \\
        \mathbf{s}_{ji} & \mathbf{s}_{ji}^{\times}
    \end{bmatrix}\mathbf{x} -  \begin{bmatrix}
        0 & -\mathbf{o}{ji}^T \\
        \mathbf{o}_{ji} & -\mathbf{o}_{ji}^{\times}
    \end{bmatrix} \mathbf{x} = \mathbf{0} \\
    \begin{bmatrix}
        0 & -(\mathbf{s}_{ji} - \mathbf{o}_{ij})^T \\
        (\mathbf{s}_{ji} - \mathbf{o}_{ji}) & (\mathbf{s}_j + \mathbf{s}_i + \mathbf{o}_j + \mathbf{o}_i)^{\times}
        \end{bmatrix}_{4 \times 4} \mathbf{x} &= \mathbf{0} 
        \label{eq:expected_measurement}
\end{align}
Note that $\mathbf{x}$ lies in the null space of $\mathbf{H}_t$.
Similar to~\cite{srivatsan2016estimating}, a \textit{pseudo measurement model} for the Kalman filter is defined:
\begin{align}
    \mathbf{H}_t \mathbf{x} &= \mathbf{z}^h \quad .
    \label{eq:measurement_model}
\end{align}
Optimal alignment of translation invariant measurements $\mathbf{s}_{ji}$ and $\mathbf{o}_{ji}$ is given by the state $\mathbf{x}$, that minimizes Equation \eqref{eq:measurement_model}. We force the pseudo measurement model $\mathbf{z}^h=0$. The pseudo measurements is associated with uncertainties that depend on $\mathbf{x}_t, \mathbf{s}_{ji}$ and $\mathbf{o}_{ji}$.
We assume that $\mathbf{x}$ and $\mathbf{z}_t$ are Gaussian distributed. Subsequently, by considering a static process model, the Kalman equations are given by
\begin{align}
    \mathbf{x}_{t} &= \bar{\mathbf{x}}_{t-1} - \mathbf{K}_t \left( \mathbf{H}_t \bar{\mathbf{x}}_{t-1} \right) \\
    \Sigma^{\mathbf{x}}_{t} &= \left( \mathbf{I} - \mathbf{K}_t \mathbf{H}_t \right) \bar{\Sigma}^{\mathbf{x}}_{t-1} \\
    \mathbf{K}_t &= \bar{\Sigma}^\mathbf{x}_{t-1} \mathbf{H}_t^T \left( \mathbf{H}_t\bar{\Sigma}^\mathbf{x}_{t-1} \mathbf{H}_t^T + \Sigma_t^{\mathbf{h}}\right)^{-1} \quad , \label{eq:kalman_equations}
\end{align}
where $\bar{\mathbf{x}}_{t-1}$ is the normalized mean of the state estimate at $t-1$, $\mathbf{K}_t$ is the Kalman gain and $\bar{\Sigma}^{\mathbf{x}}_{t-1}$ is the covariance matrix of the state at $t-1$. The parameter $\Sigma_t^{\mathbf{h}}$ is the measurement uncertainty at timestep $t$ which is state-dependent and is defined as follows~\cite{choukroun2006novel}:
\begin{align}
    \Sigma_t^{\mathbf{h}} = \frac{1}{4}\rho\left[ tr(\bar{\mathbf{x}}_{t-1}\bar{\mathbf{x}}_{t-1}^T + \bar{\Sigma}^{x}_{t-1})\mathbb{I}_4 - (\bar{\mathbf{x}}_{t-1}\bar{\mathbf{x}}_{t-1}^T + \bar{\Sigma}^{x}_{t-1} )\right] \quad ,
    \label{eq:choukron}
\end{align}
where $\rho$ is a constant which corresponds to the uncertainty of the correspondence measurements and is set empirically.
Since we have multiple measurements, we can incorporate all simultaneously into the filter by introducing $\mathbf{G}_t$ as 
\begin{equation}
    \mathbf{G}_t = \left[ \mathbf{H}_1, \ldots, \mathbf{H}_{N_c}\right] \in \mathbb{R}^{N_c \times 4} \quad ,
\end{equation}
where $N_c$ is the number of translation invariant measurements obtained so far. 
% Matrix $\mathbf{H}_t$ is replaced by $\mathbf{G}_t$ in Equation \eqref{eq:kalman_equations} for the case of multiple measurements.
In order for the state to represent a rotation, a common technique is used to normalize the state after a prediction step as
\begin{equation}
    \bar{\mathbf{x}}_{t} = \frac{\mathbf{x}_{t}}{||\mathbf{x}_{t}||_2} \quad \bar{\Sigma}^{\mathbf{x}}_{t} = \frac{\Sigma^{\mathbf{x}}_{t}}{||\mathbf{x}_{t}||_2^2} \quad .
\end{equation}
Once the rotation is estimated using the Kalman Filter, computing the translation from~\eqref{eq:generativemodel} as:
\begin{equation}
    \mathbf{t} = \mathbf{s}_i - \mathbf{R}\mathbf{o}_i \quad ,
\end{equation}
where $\mathbf{R} \in SO(3)$ form of the quaternion output $\mathbf{x}_t$ from the Kalman filter.
However, it is useful to use the centroids of the points instead of one correspondence pair to handle noisy measurements as:
\begin{equation}
    \mathbf{t} = \frac{\sum_{i=0}^{N} \mathbf{s}_i}{N} - \mathbf{R}\frac{\sum_{i=0}^{N} \mathbf{o}_i}{N} \quad .
\end{equation}

With each iteration of the update step of the Kalman filter, we obtain a new homogeneous transformation ${}^{W} H_{\mathcal{F}}$ which is then used to transform the model. The transformed model is used to recompute correspondences and repeat the Kalman Filter update steps. Similar to ICP, we calculate the change in homogeneous transformation between iterations and/or maximum number of iterations in order to check for convergence. We denote the change as $ {}_t\Delta_{LKF} = \left[ RMSE(\bar{\mathbf{x}}_t, \bar{\mathbf{x}}_{t-1}), RMSE(\mathbf{t}_t, \mathbf{t}_{t-1})\right]^T$ and $\epsilon = [\epsilon_x, \epsilon_t]^T$ as the corresponding convergence threshold. 
% In practice, since there is a possibility to get stuck in an infinite loop, we also use a pre-defined maximum number of iterations as a convergence criterion.

%  We arrived at a truly linear Kalman filter, that is dependent on the chosen action $a_t$.
% \todo[inline]{Add dependence on a in H, i.e. H(at)}
% \todo[inline]{Is sigma h the measurement noise from z? If so then make a gaussian distribution }
% \todo[inline]{Verify equations}

\input{figures/pseudo_code}

 %%%%%%%%%%%%%%%%%%%%%%%%%%%%%%Active Touch Selection%%%%%%%%%%%%%%%%%%%%%%%%%%
\subsection{Active Touch Selection}
\label{sec:active_selection}
To reduce the number of touches required to converge to the true position of the object, we need to make an informed decision on which action $\mathbf{a}_{t}$ to perform next based on the current state estimate. The set of possible actions is constrained by the position of the object in the workspace and the reachability of that position by the robot, and we define it as $\mathcal{A}$. We generate the set of actions $\mathcal{A}$ by sampling uniformly along the faces of a bounding box on the current estimate of the object pose. We define an action as a ray represented by a tuple $\mathbf{a} = (\mathbf{n}, \mathbf{d})$, with $\mathbf{n}$ as the start point and $\mathbf{d}$ the direction of the ray. We seek to choose the action $\mathbf{a}^{*}_{t}$, that \textit{maximizes} the overall \textit{Information Gain}. We measure the information gain as the Kullback–Leibler (KL) divergence between the posterior distribution $p(\mathbf{x} | \mathbf{z}_{1:t}, \mathbf{a}_{1:t})$ after executing action $\mathbf{a}_{t}$ and the prior distribution $p(\mathbf{x} | \mathbf{z}_{1:t-1}, \mathbf{a}_{1:t-1})$.
However, it must be noted that the future measurements $\mathbf{z}_{t}$ are hypothetical. Similar to~\cite{saund2017touch}, we approximate our action-measurement model $p(\mathbf{z}_{t} | \mathbf{x}, \mathbf{a}_{t})$ as a ray-mesh intersection in simulation to extract the hypothetical measurement given a certain action when the object is at the estimated pose.
% We use the Möller–Trumbore intersection algorithm~\cite{moller1997fast} in order to perform the ray-mesh intersection to extract hypothetical measurements. Furthermore, we parallelize the ray-mesh intersection in order to speed up computation. 
For each hypothetical action $\hat{\mathbf{a}}_t \in \mathcal{A}(\mathbf{x}_{t-1})$ and the hypothetical measurement $\hat{\mathbf{z}}_{t}$, we estimate the posterior by $p(\mathbf{x} | \hat{\mathbf{z}}_{1:t}, \hat{\mathbf{a}}_{1:t})$ known as the \textit{one-step look ahead}. 
Therefore we perform the most optimal action $\mathbf{a}_{t}^*$ with the robot given by
\begin{align}
    &\mathbf{a}_{t}^* 
    % \argmax_{\hat{\mathbf{a}}_{t}} KL(p(\mathbf{x} | \hat{\mathbf{z}}_{1:t},  \hat{\mathbf{a}}_{1:t}) || p(\mathbf{x} | \mathbf{z}_{1:t-1}, \mathbf{a}_{1:t-1})) \\
    &= \argmax_{\hat{\mathbf{a}}_{t}} \int_{\mathbf{x}} p(\mathbf{x} | \hat{\mathbf{z}}_{1:t},  \hat{\mathbf{a}}_{1:t}) \log\frac{p(\mathbf{x} | \hat{\mathbf{z}}_{1:t},  \hat{\mathbf{a}}_{1:t})}{p(\mathbf{x} | \mathbf{z}_{1:t-1}, \mathbf{a}_{1:t-1})}d \mathbf{x} \quad . \label{eq:kl_div_prio_post}
\end{align}
Given that the prior and posterior are multivariate Gaussian distributions, the KL divergence in \eqref{eq:kl_div_prio_post} can be computed in closed form as~\cite{duchi2007derivations}:
\begin{align}
    \mathbf{a}_{t}^* = \argmax_{\hat{a}_{t}} \frac{1}{2} &\left[ log\frac{det(\bar{\Sigma}_{t-1})}{det(\hat{\bar{\Sigma}}_{t})} + Tr(\bar{\Sigma}_{t-1}^{-1} \hat{\bar{\Sigma}}_{t})) - d \right. \nonumber \\ 
    &\left. + (\hat{\bar{\mathbf{x}}}_{t} - \bar{\mathbf{x}}_{t-1})^T \bar{\Sigma}_{t}^{-1} (\hat{\bar{\mathbf{x}}}_{t} - \bar{\mathbf{x}}_{t-1}) \vphantom{log\frac{det(\Sigma_{t})}{det(\hat{\Sigma}_{t-1})}}\right] \quad ,
    \label{eq:kld_closed_form}
\end{align}
where $d$ is the dimension of the state vector and $d=4$ in our case.
This enables us to evaluate an exhaustive list of actions at marginal computation cost in \textit{real time} without the need to prune actions or setting trade-offs with computation time as compared to prior work. When the stop criterion ${}_{t|t}\Delta_{LKF}\leq \xi$, which is defined similar to the convergence criterion ${}_t\Delta_{LKF}$, is reached, no further actions are performed.
% This is defined similarly to the convergence criterion ${}_t\Delta_{LKF}$ from the previous section with the only difference being that the stop criterion will compare the change before and after each action. 
The overall algorithm is shown in Algorithm 1.

%\begin{algorithm}[t!]
%%\SetAlgoLined
%\KwResult{$\mathbf{q}$, $\mathbf{t}$, $\Sigma^{q}$}
%\textbf{Input: }{$\mathbf{q_0}$, $\mathbf{t_0}$, $\epsilon_q$, $\epsilon_t$, $\mathbf{q_{gt}}$, %$\mathbf{t_{gt}}$} \\ 
%\textbf{Initialisation:} \\
% 
%%\caption{Translation-Invariant Quaternion Filter}
%\end{algorithm}

%% file: figures/pseudo_code.tex
\begin{algorithm}[t!]
\SetAlgoLined
\textbf{Input: }${}^v \mathbf{x}$, ${}^v \Sigma^{x}$, ${}^v \mathbf{t}$, $\mathcal{O}$ \\ 
\KwResult{${}^t \mathbf{x}$, ${}^t \Sigma^{x}$, ${}^t \mathbf{t}$} 
\textbf{Initialisation:} \\
$\mathbf{x}_t \leftarrow {}^v \mathbf{x}$, $\Sigma^{x}_t \leftarrow {}^v \Sigma^{x}$, $\mathbf{t}_t \leftarrow {}^v \mathbf{t}$ \\
Measurements ${}^t \mathcal{S} \leftarrow$ \{\}, Correspondences $\mathcal{C}\leftarrow$ \{\}, \\
Actions $\mathcal{A} \leftarrow$ \{\}, Sim. Measurements $\mathcal{Z}\leftarrow$ \{\}, \\
KL Divergence $\mathcal{D}_{KL} \leftarrow$ \{\} \;
%Transform $\mathcal{M}$ to $q_0, t_0$ \\
%Estimated Pose $\mathbf{\hat{q}}, \mathbf{t} \leftarrow \mathrm{MAX\_VALUE}$ \\
%Initial Error $\Delta \mathbf{q} = abs(Eul(\mathbf{\hat{q}}) - Eul(\mathbf{q}))$, $\Delta \mathbf{t} = abs(Eul(\mathbf{\hat{q}}) - Eul(\mathbf{q}))$ \\ 
\While{($\Delta \mathbf{x} > \xi_x$ and $\Delta \mathbf{t} > \xi_t$)}{
    
    $\hat{\mathcal{O}}\leftarrow$ transform($\mathcal{O}$, $\mathbf{x}_t$, $\mathbf{t}_t$) \;
%    min, max $\leftarrow$ create\_bounding\_box($\hat{\mathcal{O}}$) \;
    \eIf{size(${}^t \mathcal{S}$)$\leq2$}{
        $\mathbf{a}^*_t$ = select\_random\_action($\hat{\mathcal{O}}$) \;
    }
    {
        $\mathcal{A} \leftarrow$ generate\_possible\_actions($\hat{\mathcal{O}}$) \;
        $\mathcal{Z} \leftarrow$ simulate\_measurements($\mathcal{A}, \hat{\mathcal{O}}$) \;
        $\mathcal{D}_{KL} \leftarrow$ \{\} \;
        \For{$\hat{\mathbf{z}}_t$ in $\mathcal{Z}$}{
        
            ${}^t \hat{\mathcal{S}} \leftarrow {}^t \mathcal{S} \cup \{ \hat{\mathbf{z}}_t\}$ \;
            $\hat{\mathcal{C}}$ = estimate\_correspondences($\hat{\mathcal{O}}$, ${}^t \hat{\mathcal{S}}$) \;
            $\hat{\mathbf{x}}_t$, $\hat{\Sigma}_t^{\mathbf{x}}\leftarrow$ update\_TIQF($\mathbf{x}_t$, $\Sigma^{\mathbf{x}}_t$, $\hat{\mathcal{C}}$) \;
            $KL\leftarrow$ compute\_kl\_div($\mathbf{x}_t$, $\Sigma^{\mathbf{x}}_t$, $\hat{\mathbf{x}}_t$, $\hat{\Sigma}_t^{\mathbf{x}}$) \;
            $\mathcal{D}_{KL}\leftarrow \mathcal{D}_{KL} \cup KL$ \;
        
        }
        $\mathbf{a}^*_t \leftarrow$ choose\_best\_action($\mathcal{A}$, $\mathcal{D}_{KL}$) \;
        $\mathbf{z}_t \leftarrow$ execute\_action($\mathbf{a}^*_t$) \;
        ${}^t \mathcal{S} \leftarrow {}^t \mathcal{S} \cup \{ \mathbf{z}_t \}$ \;
        $\mathcal{C}$ = estimate\_correspondences($\hat{\mathcal{O}}$, ${}^t \mathcal{S}$) \;
        $\mathbf{x}_t$, $\Sigma^{x}_t \leftarrow$ update\_TIQF($\mathbf{x}_t$, $\Sigma^{\mathbf{x}}_t$, $\mathcal{C}$) \;
        $\mathbf{t}_t \leftarrow$ compute\_translation($\mathbf{x}_t$, $\mathcal{C}$) \;
        %$\mathbf{a}$ = $a_L$[argmax(KL)] \;
     }
}   
\label{algo:pseudo_code}
\caption{Active touch for tactile point cloud registration and accurate object localization}
\end{algorithm}
%% remove algo

%% file: sections/experiments.tex
\section{EXPERIMENTS}
\label{sec:experiment}

The experimental setup shown in Figure~\ref{fig:real_robot} consists of Universal Robots UR5 robot with a Robotiq 2F140 Gripper. The robot is attached to a specially designed pedestal and mounted at $135^{o}$ with respect to the vertical axis.
%overlooking the workspace wherein the test objects are placed.
The standard gripper pads of the Robotiq 2F140 are replaced with tactile sensors from XELA Robotics\footnote{https://xelarobotics.com/} on the fingertips and the phalanges as shown in Figure~\ref{fig:real_robot}. The tactile sensing system consists of $N_T = 140$ taxels that provide 3-axis force measurements on each taxel in the sensor coordinate frame. It is composed of eight tactile sensors in total, where 4 tactile sensors are on each finger: phalange sensor (24 taxels), outer finger (24 taxels), finger tip (6 taxels) and inner finger (16 taxels).
The tactile sensors function on the principle of Hall-effect sensing and are covered with a soft, textile material~\cite{tomo2019development}. A contact is established with the object when the norm of the 3-axis force value of any taxel $f_r$ exceeds a threshold $\tau_f$, which is defined with respect to the baseline values and has been tuned empirically. The 3D positions of the contacted taxels are transformed into the robot base frame using robot kinematics and are appended to the tactile point cloud ${}^t\mathcal{S}$. 
An Azure Kinect DK RGB-D camera is placed in front of the workspace, which provides the vision point cloud ${}^v\mathcal{S}$.
Simulation and experimental results are provided in the following sections.
All simulation and real experiments were executed on a workstation running Ubuntu 18.04 with 8 core Intel i7-8550U CPU @ 1.80GHz and 16 GB RAM.
% The experimental setup is associated with distinct coordinate frames as shown in Figure~\ref{fig:expsetup} wherein $\mathcal{W}$ refers to the world frame, $\mathcal{R}$ is the robot-base frame, $\mathcal{C}$ is the camera frame, $\mathcal{E}$ is the robot end-effector frame, $\mathcal{T}_{ij}$ are the tactile sensor taxel frames where $i = 1, 2, 3, \dots, 140$ and $j=1,2,\dots, 8$. $\mathcal{O}_k$ is the object frames which are rigidly attached to each object ($k = 1,2,3, \dots, m$). The problem is to find ${}^{\mathcal{W}}H_{\mathcal{O}_k}$ for each object $k$ in the workspace. 

%%%%%%%%%%%%%%%%%%% Experimental Object %%%%%%%%%%%%
%\input{figures/object_table}

% \begin{figure}[t!]
% \centering
%   \includegraphics[width =.85 \columnwidth]{figures/ExperimentalObjects-Final.png}
%   \caption{Daily objects used for robot experiments}
%     \label{fig:object_list}
% \end{figure}
%%%%%%%%%%%%%%%%%%%%%%%%%%%%%%%%%%%%%%%%%%%%%%%%%%%%

%%%%%%%%%%%%%%%% Plots%%%%%%%%%%%%%%%%%%%%

%%%%%%%%%%%%%%%%%%%%%%%%%%%%%%%%%%%%%%%%
%\begin{figure}[H]
%    \centering
%    \includegraphics[width = \columnwidth]{figures/experiment_setup.png}
%    \caption{Experimental Setup}
%    \label{fig:expsetup}
%\end{figure}

% \begin{figure}[t!]
%     \centering
%     \includegraphics[width = \columnwidth]{figures/bunny_sim.pdf}
%     \caption{Active touch point selection and pose estimation in simulation}
%     \label{fig:bunny_sim }
% \end{figure}

\subsection{Simulation Results}
In order to validate and compare our method to the state of the art~\cite{arun2019registration} which we use as baseline, we perform extensive simulation experiments using the Stanford 3D Scanning Repository~\cite{stanford}. We assume unknown correspondences to correspond to realistic scenarios. 
% Point correspondences are found using the closest point rule~\cite{besl1992method}.
We added noise that is sampled randomnly from a normal distribution $\mathcal{N}(0, 5\times10^{-3})$ to the cloud obtained from the meshes, henceforth called \textit{scene}. We set the initial start pose for each model sampled uniformly from $[-50, 50]mm$ and $[-30^o, 30^o]$ for position and orientation respectively. The initial state $\mathbf{x}_0$ is obtained from the initial start pose and the initial covariance $\Sigma^{\mathbf{x}}_0$ is set to $\mathbb{I}_4$. In order to simulate tactile measurements, we sequentially sample points from the \textit{scene} and register to the model cloud using our TIQF estimation. We compare random sampling versus active sampling of points. We repeated each experiment 100 times for each model.
% In the case of random sampling, a bounding box for the \textit{scene} is extracted based on the initial estimation of the \textit{scene}.
Actions are uniformly sampled on each face of the bounding box encapsulating the scene and we use ray-mesh (triangle) intersection algorithm in order to extract the measured points. We use the Möller–Trumbore intersection algorithm~\cite{moller1997fast} in order to perform the ray-mesh intersection to extract hypothetical measurements.
For random action selection, an action is randomnly selected from the sampled set of actions and is executed.
For active touch selection, hypothetical measurements $\hat{\mathbf{z}}$ are extracted using the generated actions and one-step lookahead for each action-hypothetical measurement pair is performed by running the TIQF algorithm for a fixed number of iterations. The optimal action $\mathbf{a}^*_t$ is chosen which is associated with the largest KL divergence of the hypothetical posterior with the prior belief.
For all the models in simulation, we generate a total of 100 possible actions in order to choose the optimal action at each measurement step. Furthermore, it was noted that due the low number of sparse points available for registration, the TIQF algorithm often gets into local minima. To tackle this, we employ a well known strategy to add local perturbations sampled from an uniform distribution $[-2^o, 2^o]$ around the local minima. 
We report the simulation results showing the root mean square error (RMSE) of translation and rotation versus number of points in Figure~\ref{fig:sim_random_active}. 

%\input{figures/sim_random_v_active}

% \todo[inline]{Explain why the starting point is different}

%%%%%%%%%%%%%%%%%%%%%%%%%%%%%%%%%%%%%%%%%%%%%
\begin{table}[b!]
\centering
\resizebox{\columnwidth}{!}{%
\begin{tabular}{|c|c|c|}
\hline
\textbf{\# Actions} & \textbf{Simulated Mesh (s)} & \textbf{Real Object Mesh (s)} \\ \hline
\hline
10 & 0.33 & 0.17 \\ \hline
100 & 5.06 &  1.75\\ \hline
1000 & 42.56 & 13.70 \\ \hline
\end{tabular}%
}
\caption{The computation time required for action generation and action selection with the one-step look ahead for simulated mesh ($\approx 5000$ triangular faces) and real object mesh ($1000$ triangular faces). The performance shown here is representative as it is dependent on chosen hardware.}
\label{tab:action-table}
\end{table}
%%%%%%%%%%%%%%%%%%%%%%%%%%%%%%%%%%%%%%%%%%%%%

\input{figures/sim_random_v_active}

\subsection{Robot Experimental Results}
% \todo[inline]{make fig in two col}
In order to validate our proposed framework with robotic systems, we chose 6 daily objects of various intrinsic properties as shown in Figure~\ref{fig:real_robot}. We used the following objects: shampoo, sugar box, spray, cleaner, can, and olive oil bottle. The objects have been chosen according to the following criteria: varying shape between simple (cuboid, cylinder) to complex (for instance, spray) and varying degrees of transparency (for instance, highly transparent cleaner, highly opaque sugar box). The corresponding CAD meshes for the real objects were obtained using a high precision 3D scanner. The objects are rigidly attached to the workspace and the ground truth is extracted with respect to the world frame $\mathcal{W}$. The objects are moved randomly around the workspace between experiments to evaluate the robustness. 
% We evaluate our active methodology on the real objects.
% \paragraph{Action Selection and Execution}
% write about motion planning
Since the pose estimate of the objects are unknown, the robot actions are performed as \textit{guarded motions} so that the robots do not topple the other objects in the workspace~\cite{petrovskaya2011global}. The initial estimate is computed from the vision point cloud by using the TIQF estimation as described in the previous section. The actions are generated uniformly with directions along coordinate axes on the 5 faces of the bounding box around the current estimate, assuming that it is unfeasible to contact the object from the bottom when placed on a table.
The action list is pruned in order to remove actions that are kinematically unfeasible and that collide with the workspace. 
Hypothetical actions-measurement pairs are generated with the ray-mesh intersection with the current estimate of the object. 
The candidate action with the highest expected information gain with one-step lookahead is chosen and performed on the real object. As the actions may not contact the real objects as they are based on the current estimate, we note that \textit{negative information} i.e., information about absence instead of presence of measurements is not considered. However, since the action generation and selection is guided initially by the vision estimate and iteratively updated with the tactile measurements, empirically we find fewer actions resulting in negative information.
% Upon selection of the candidate measurement action $a^{*}_t$, the robot moves
% to start point of the action $\mathbf{n}_t$ and move along $\mathbf{d}_t$ until contact is detected. 
%We employ simple linear cartesian motions for all associated robot motions.
Our motion planner ensures the robot safely moves to start positions of actions by moving over the workspace $W_{XYZ}$ at a height larger than the biggest object and descends vertically to the start point of the selected action. Furthermore, in order to prevent the estimation to get stuck in local minima, similar to the simulation experiments we add local perturbations to the object at each iteration.
We performed localisation trials for the 6 objects and repeated the trials ten times on each object by varying their ground truth locations. The results of the experiments is presented in the Figure~\ref{fig:real_random_active}. 

\input{figures/real_sim_random_v_active}
\subsection{Discussion}
Figure~\ref{fig:sim_random_active} shows that across all simulated object, our proposed active strategy outperforms random strategy used by existing state-of-the-art methods in terms of accuracy (average RMSE for rotation and translation) and convergence rate with respect to number of points for both rotation and translation. Furthermore, the experimental results show that in first 5 measurements, the RMSE for translation and rotation for our proposed strategy is markedly lower than random approach. This demonstrates that our proposed method performs effectively right from the first touch. The results in simulation are corroborated with the experiments with the selected daily objects as seen in Figure~\ref{fig:real_random_active}. Moreover, as we intentionally chose objects with varying degree of transparency such as the cleaner that causes issues for vision sensor but is accurately localised with tactile sensing. As noted earlier, our proposed method can reason over multiple candidate actions to find the most optimal action using very low computation time without the need for high compute hardware. This is shown in Table~\ref{tab:action-table} for a simulated mesh and a real object mesh. On the other hand, for objects with an axis of symmetry such as the "Can" in Figure~\ref{fig:real_robot}, there are infinite solutions for rotation estimation. As our proposed method estimates rotation and subsequently computes translation, we effectively cannot compute the RMSE for rotation and translation for such symmetric objects as similarly noted by~\cite{gross2019alignnet}. As part of future work, we will investigate solutions for tackling the pose estimation of symmetric objects by incorporating texture and colour in the point clouds.

%% file: figures/sim_random_v_active.tex
\newcommand\height{7cm}
\newcommand\width{\columnwidth}
%\pgfplotsset{every axis title/.append style={at={(0.1,0.8)}}}
\begin{figure}[t]
    \centering
    \begin{subfigure}[b]{\columnwidth}
         \centering
             \begin{tikzpicture}
                \begin{groupplot}[group style={group size= 1 by 2, 
                x descriptions at=edge bottom, 
%                y descriptions at=edge left, 
                ylabels at=edge left, 
                xlabels at=edge bottom,
                horizontal sep=0.3cm,
                vertical sep=0.1cm
                },
                xmajorgrids=true,
                ymajorgrids=true,
                grid style={line width=.1pt, draw=gray!10},major grid style={line width=.2pt,draw=gray!50},
                xtick pos=bottom,
                ytick pos=left,
                xmin=3,
                xmax=20,
                ylabel shift = -5pt,
                xlabel shift = -5pt,
                height=0.25*\height,
                scale only axis,
                width=5cm,
                xlabel={\# points},
                tick label style={font=\scriptsize},
                title style={yshift=-1.3ex, font=\scriptsize},
                legend style={nodes={scale=0.4, transform shape}},
                y label style={at={(0,0.5)}, font=\scriptsize, yshift=+2.0ex},
                label style={font=\scriptsize}
                ]
            
            % ================================LOADING=========================================
            %-- Load rotation
            %Random
            \pgfplotstableread[col sep=semicolon]{data/sim_random_v_active/overall_rotation_random.csv}\simoverallrandnomrot;
            %Active
            \pgfplotstableread[col sep=semicolon]{data/sim_random_v_active/overall_rotation_active.csv}\simoverallactiverot;
            
            %-- Load translation
            %Random
            \pgfplotstableread[col sep=semicolon]{data/sim_random_v_active/overall_translation_random.csv}\simoverallrandomtrans;
            %Active
            \pgfplotstableread[col sep=semicolon]{data/sim_random_v_active/overall_translation_active.csv}\simoverallactivetrans;

            \nextgroupplot[ylabel=\tiny{RMSE Rot. $[deg]$}, title=Average, title style={text depth = 0pt}]
                \addplot [color=green!60!black, line width=1.5pt, mark=none, dotted] table [x=x, y=mean, col sep=semicolon, row sep=newline, /pgf/number format/read comma as period]{\simoverallrandnomrot};
                \addplot [name path=r_down, color=green!60!black!50, draw opacity=0.5, forget plot] table [x=x, y=lmean, col sep=semicolon, row sep=newline, /pgf/number format/read comma as period]{\simoverallrandnomrot};
                \addplot [name path=r_top, color=green!60!black!50, draw opacity=0.5, forget plot] table [x=x, y=umean, col sep=semicolon, row sep=newline, /pgf/number format/read comma as period]{\simoverallrandnomrot};
                \addplot[green!60!black!50,fill opacity=0.5, forget plot] fill between[of=r_top and r_down];
                
                \addplot [color=red, line width=1.5pt] table [x=x, y=mean, col sep=semicolon, row sep=newline, /pgf/number format/read comma as period]{\simoverallactiverot};
                \addplot [name path=a_down, color=red!50, draw opacity=0.5, forget plot] table [x=x, y=lmean, col sep=semicolon, row sep=newline, /pgf/number format/read comma as period]{\simoverallactiverot};
                \addplot [name path=a_top, color=red!50, draw opacity=0.5, forget plot] table [x=x, y=umean, col sep=semicolon, row sep=newline, /pgf/number format/read comma as period]{\simoverallactiverot};
                \addplot[red!50,fill opacity=0.5, forget plot] fill between[of=a_top and a_down];
                
                \draw[black, thin, densely dotted] (axis cs:\pgfkeysvalueof{/pgfplots/xmin},0) -- (axis cs:\pgfkeysvalueof{/pgfplots/xmax},0);
                \legend{Random, Active}
                % \addplot table [x=x, y=Active, col sep=semicolon, row sep=newline, /pgf/number format/read comma as period]{\mydata};
                % \legend{$Random$,$Active$}
                
            \nextgroupplot[ylabel=\tiny{RMSE Trans. $[cm]$}]
                \addplot [color=green!60!black!60!black, line width=1.5pt, , mark=none, dotted] table [x=x, y=mean, col sep=semicolon, row sep=newline, /pgf/number format/read comma as period]{\simoverallrandomtrans};
                \addplot [name path=r_down, color=green!60!black!60!black!50, draw opacity=0.5, forget plot] table [x=x, y=lmean, col sep=semicolon, row sep=newline, /pgf/number format/read comma as period]{\simoverallrandomtrans};
                \addplot [name path=r_top, color=green!60!black!60!black!50, draw opacity=0.5, forget plot] table [x=x, y=umean, col sep=semicolon, row sep=newline, /pgf/number format/read comma as period]{\simoverallrandomtrans};
                \addplot[green!60!black!60!black!50,fill opacity=0.5, forget plot] fill between[of=r_top and r_down];
                
                \addplot [color=red, line width=1.5pt] table [x=x, y=mean, col sep=semicolon, row sep=newline, /pgf/number format/read comma as period]{\simoverallactivetrans};
                \addplot [name path=a_down, color=red!50, draw opacity=0.5, forget plot] table [x=x, y=lmean, col sep=semicolon, row sep=newline, /pgf/number format/read comma as period]{\simoverallactivetrans};
                \addplot [name path=a_top, color=red!50, draw opacity=0.5, forget plot] table [x=x, y=umean, col sep=semicolon, row sep=newline, /pgf/number format/read comma as period]{\simoverallactivetrans};
                \addplot[red!50,fill opacity=0.5, forget plot] fill between[of=a_top and a_down];
                
                \draw[black, thin, densely dotted] (axis cs:\pgfkeysvalueof{/pgfplots/xmin},0) -- (axis cs:\pgfkeysvalueof{/pgfplots/xmax},0);
                \legend{Random, Active}
                % \addplot table [x=x, y=Active, col sep=semicolon, row sep=newline, /pgf/number format/read comma as period]{\mydata};
                % \legend{$Random$,$Active$}
            \end{groupplot}
            \end{tikzpicture}

     \end{subfigure}
     \hfill
     \begin{subfigure}[b]{\columnwidth}
         \centering
         \begin{tikzpicture}
            \begin{groupplot}[group style={group size= 5 by 2,
                x descriptions at=edge bottom, 
%                y descriptions at=edge left, 
                ylabels at=edge left, 
                xlabels at=edge bottom,
                horizontal sep=0.3cm,
                vertical sep=0.1cm
                },
                xmajorgrids=true,
                ymajorgrids=true,
                grid style={line width=.1pt, draw=gray!10},major grid style={line width=.2pt,draw=gray!50},
                xtick pos=bottom,
                ytick pos=left,
                xmin=3,
                xmax=20,
                ylabel shift = -5pt,
                xlabel shift = -5pt,
                height=0.15*\height,
                scale only axis,
                width=1.3cm,
                xlabel={\# points},
                tick label style={font=\tiny},
                title style={yshift=-1.3ex, font=\tiny},
                label style={font=\tiny}
                ]
                
                % ================================LOADING=========================================
                %-- Loading bunny
                % Rotation
                \pgfplotstableread[col sep=semicolon]{data/sim_random_v_active/bunny_random_rotation.csv}\simbunnyrandomrotation;
                \pgfplotstableread[col sep=semicolon]{data/sim_random_v_active/bunny_active_rotation.csv}\simbunnyactiverotation;
                % Translation
                \pgfplotstableread[col sep=semicolon]{data/sim_random_v_active/bunny_random_translation.csv}\simbunnyrandomtranslation;
                \pgfplotstableread[col sep=semicolon]{data/sim_random_v_active/bunny_active_translation.csv}\simbunnyactivetranslation;

                %-- Loading Buddah
                %Rotation
                \pgfplotstableread[col sep=semicolon]{data/sim_random_v_active/buddah_random_rotation.csv}\simbuddahrandomrotation;
                \pgfplotstableread[col sep=semicolon]{data/sim_random_v_active/buddah_active_rotation.csv}\simbuddahactiverotation;
                %Translation
                \pgfplotstableread[col sep=semicolon]{data/sim_random_v_active/buddah_random_translation.csv}\simbuddahrandomtranslation;
                \pgfplotstableread[col sep=semicolon]{data/sim_random_v_active/buddah_active_translation.csv}\simbuddahactivetranslation;

                %-- Loading Dragon
                %Rotation
                \pgfplotstableread[col sep=semicolon]{data/sim_random_v_active/dragon_random_rotation.csv}\simdragonrandomrotation;
                \pgfplotstableread[col sep=semicolon]{data/sim_random_v_active/dragon_active_rotation.csv}\simdragonactiverotation;
                %Translation
                \pgfplotstableread[col sep=semicolon]{data/sim_random_v_active/dragon_random_translation.csv}\simdragonrandomtranslation;
                \pgfplotstableread[col sep=semicolon]{data/sim_random_v_active/dragon_active_translation.csv}\simdragonactivetranslation;

                %-- Loading Armadillo
                %Rotation
                \pgfplotstableread[col sep=semicolon]{data/sim_random_v_active/armadillo_random_rotation.csv}\simarmadillorandomrotation;
                \pgfplotstableread[col sep=semicolon]{data/sim_random_v_active/armadillo_active_rotation.csv}\simarmadilloactiverotation;        
                %Translation
                \pgfplotstableread[col sep=semicolon]{data/sim_random_v_active/armadillo_random_translation.csv}\simarmadillorandomtranslation;
                \pgfplotstableread[col sep=semicolon]{data/sim_random_v_active/armadillo_active_translation.csv}\simarmadilloactivetranslation;

                %-- Loading Lucy
                %Rotation
                \pgfplotstableread[col sep=semicolon]{data/sim_random_v_active/lucy_random_rotation.csv}\simlucyrandomrotation;
                \pgfplotstableread[col sep=semicolon]{data/sim_random_v_active/lucy_active_rotation.csv}\simlucyactiverotation;
                %Translation
                \pgfplotstableread[col sep=semicolon]{data/sim_random_v_active/lucy_random_translation.csv}\simlucyrandomtranslation;
                \pgfplotstableread[col sep=semicolon]{data/sim_random_v_active/lucy_active_translation.csv}\simlucyactivetranslation;

                %horizontal sep=1mm,
                %vertical sep=1mm},
                % =========================================================================

                % =====================================Rotation=====================================
                % Plot bunny
                \nextgroupplot[ylabel=\tiny{rmse (${}^o$)}, y label style={at={(0,0.5)}, font=\tiny, yshift=+1.5ex}, title=Bunny, title style={text depth = 0pt}]
                    %\node [draw=black] at (rel axis cs:0.8, 0.85){\includegraphics[width=0.8cm, height=0.8cm]{figures/bunny_crop.png}};
                    %\node [draw=black] at (rel axis cs:0.79, 0.81){\includegraphics[width=0.8cm, height=1cm]{figures/bunny_crop.png}};
                    % Plot Random
                    \addplot [color=green!60!black, line width=1.5pt, mark=none, dotted] table [x=x, y=mean, col sep=semicolon, row sep=newline, /pgf/number format/read comma as period]{\simbunnyrandomrotation};
                    % Plot Active
                    \addplot [color=red, line width=1.5pt] table [x=x, y=mean, col sep=semicolon, row sep=newline, /pgf/number format/read comma as period]{\simbunnyactiverotation};
                    
                    % Plot zero line
                    \draw[black, thin, densely dotted] (axis cs:\pgfkeysvalueof{/pgfplots/xmin},0) -- (axis cs:\pgfkeysvalueof{/pgfplots/xmax},0);
                    
                    %\node[anchor=north west, draw=black,fill=white] at (rel axis cs:0.02,0.98) {\footnotesize Bunny};
                    
                % Plot Buddha 
                \nextgroupplot[title=Buddah, title style={text depth = 0pt}]
                    % Plot Random 
                    %\node [draw=black] at (rel axis cs:0.79, 0.81){\includegraphics[width=0.8cm, height=1cm]{figures/buddah_crop.png}};
                    \addplot [color=green!60!black, line width=1.5pt, mark=none, dotted] table [x=x, y=mean, col sep=semicolon, row sep=newline, /pgf/number format/read comma as period]{\simbuddahrandomrotation};
                    % Plot Active
                    \addplot [color=red, line width=1.5pt] table [x=x, y=mean, col sep=semicolon, row sep=newline, /pgf/number format/read comma as period]{\simbuddahactiverotation};
                    % Plot zero line
                    \draw[black, thin, densely dotted] (axis cs:\pgfkeysvalueof{/pgfplots/xmin},0) -- (axis cs:\pgfkeysvalueof{/pgfplots/xmax},0);
                    
                    %\node[anchor=north west, draw=black,fill=white] at (rel axis cs:0.02,0.98) {\footnotesize Buddha};
                    
                % Plot Dragon
                \nextgroupplot[title=Dragon, title style={text depth = 0pt}]
                    %\node [draw=black] at (rel axis cs:0.78, 0.85){\includegraphics[width=1cm, height=0.8cm]{figures/dragon_crop.png}};
                    %\node [draw=black] at (rel axis cs:0.79, 0.81){\includegraphics[width=0.8cm, height=1cm]{figures/dragon_crop.png}};
                    % Plot Random 
                    \addplot [color=green!60!black, line width=1.5pt, mark=none, dotted] table [x=x, y=mean, col sep=semicolon, row sep=newline, /pgf/number format/read comma as period]{\simdragonrandomrotation};
                    % Plot Active
                    \addplot [color=red, line width=1.5pt] table [x=x, y=mean, col sep=semicolon, row sep=newline, /pgf/number format/read comma as period]{\simdragonactiverotation};
                    % Plot zero line
                    \draw[black, thin, densely dotted] (axis cs:\pgfkeysvalueof{/pgfplots/xmin},0) -- (axis cs:\pgfkeysvalueof{/pgfplots/xmax},0);
                    %\node[anchor=north west, draw=black,fill=white] at (rel axis cs:0.02,0.98) {\footnotesize Dragon};
                    
                % Plot Armadillo
                \nextgroupplot[title=Armadillo, title style={text depth = 0pt}]
                    %\node [draw=black] at (rel axis cs:0.79, 0.81){\includegraphics[width=0.8cm, height=1cm]{figures/armadillo_crop.png}};
                    % Plot Random 
                    \addplot [color=green!60!black, line width=1.5pt, mark=none, dotted] table [x=x, y=mean, col sep=semicolon, row sep=newline, /pgf/number format/read comma as period]{\simarmadillorandomrotation};
                    % Plot Active
                    \addplot [color=red, line width=1.5pt] table [x=x, y=mean, col sep=semicolon, row sep=newline, /pgf/number format/read comma as period]{\simarmadilloactiverotation};
                    % Plot zero line
                    \draw[black, thin, densely dotted] (axis cs:\pgfkeysvalueof{/pgfplots/xmin},0) -- (axis cs:\pgfkeysvalueof{/pgfplots/xmax},0);
                    
                    %\node[anchor=north west, draw=black,fill=white] at (rel axis cs:0.02,0.98) {\footnotesize Armadillo};
                    
                % Plot Lucy
                \nextgroupplot[title=Lucy, title style={text depth = 0pt}]
                    %\node [draw=black] at (rel axis cs:0.79, 0.81){\includegraphics[width=0.8cm, height=1cm]{figures/lucy_crop.png}};
                    % Plot Random 
                    \addplot [color=green!60!black, line width=1.5pt, mark=none, dotted] table [x=x, y=mean, col sep=semicolon, row sep=newline, /pgf/number format/read comma as period]{\simlucyrandomrotation};
                    % Plot Active
                    \addplot [color=red, line width=1.5pt] table [x=x, y=mean, col sep=semicolon, row sep=newline, /pgf/number format/read comma as period]{\simlucyactiverotation};
                    %\node[anchor=north west, draw=black,fill=white] at (rel axis cs:0.02,0.98) {\footnotesize Lucy};

            % =====================================Translation=====================================
                % Plot bunny
                \nextgroupplot[ylabel = \tiny{rmse (cm)}, y label style={at={(0,0.5)}, font=\tiny, yshift=+1.5ex}]
                    %\node [draw=black] at (rel axis cs:0.8, 0.85){\includegraphics[width=0.8cm, height=0.8cm]{figures/bunny_crop.png}};
                    %\node [draw=black] at (rel axis cs:0.79, 0.81){\includegraphics[width=0.8cm, height=1cm]{figures/bunny_crop.png}};
                    % Plot Random
                    \addplot [color=green!60!black, line width=1.5pt, mark=none, dotted] table [x=x, y=mean, col sep=semicolon, row sep=newline, /pgf/number format/read comma as period]{\simbunnyrandomtranslation};
                    % Plot Active
                    \addplot [color=red, line width=1.5pt] table [x=x, y=mean, col sep=semicolon, row sep=newline, /pgf/number format/read comma as period]{\simbunnyactivetranslation};
                    
                    % Plot zero line
                    \draw[black, thin, densely dotted] (axis cs:\pgfkeysvalueof{/pgfplots/xmin},0) -- (axis cs:\pgfkeysvalueof{/pgfplots/xmax},0);
                    
                    %\node[anchor=north west, draw=black,fill=white] at (rel axis cs:0.02,0.98) {\footnotesize Bunny};
                    
                % Plot Buddha 
                \nextgroupplot
                    % Plot Random 
                    %\node [draw=black] at (rel axis cs:0.79, 0.81){\includegraphics[width=0.8cm, height=1cm]{figures/buddah_crop.png}};
                    \addplot [color=green!60!black, line width=1.5pt, mark=none, dotted] table [x=x, y=mean, col sep=semicolon, row sep=newline, /pgf/number format/read comma as period]{\simbuddahrandomtranslation};
                    % Plot Active
                    \addplot [color=red, line width=1.5pt] table [x=x, y=mean, col sep=semicolon, row sep=newline, /pgf/number format/read comma as period]{\simbuddahactivetranslation};
                    % Plot zero line
                    \draw[black, thin, densely dotted] (axis cs:\pgfkeysvalueof{/pgfplots/xmin},0) -- (axis cs:\pgfkeysvalueof{/pgfplots/xmax},0);
                    
                    %\node[anchor=north west, draw=black,fill=white] at (rel axis cs:0.02,0.98) {\footnotesize Buddha};
                    
                % Plot Dragon
                \nextgroupplot
                    %\node [draw=black] at (rel axis cs:0.78, 0.85){\includegraphics[width=1cm, height=0.8cm]{figures/dragon_crop.png}};
                    %\node [draw=black] at (rel axis cs:0.79, 0.81){\includegraphics[width=0.8cm, height=1cm]{figures/dragon_crop.png}};
                    % Plot Random 
                    \addplot [color=green!60!black, line width=1.5pt, mark=none, dotted] table [x=x, y=mean, col sep=semicolon, row sep=newline, /pgf/number format/read comma as period]{\simdragonrandomtranslation};
                    % Plot Active
                    \addplot [color=red, line width=1.5pt] table [x=x, y=mean, col sep=semicolon, row sep=newline, /pgf/number format/read comma as period]{\simdragonactivetranslation};
                    % Plot zero line
                    \draw[black, thin, densely dotted] (axis cs:\pgfkeysvalueof{/pgfplots/xmin},0) -- (axis cs:\pgfkeysvalueof{/pgfplots/xmax},0);
                    %\node[anchor=north west, draw=black,fill=white] at (rel axis cs:0.02,0.98) {\footnotesize Dragon};
                    
                % Plot Armadillo
                \nextgroupplot
                    %\node [draw=black] at (rel axis cs:0.79, 0.81){\includegraphics[width=0.8cm, height=1cm]{figures/armadillo_crop.png}};
                    % Plot Random 
                    \addplot [color=green!60!black, line width=1.5pt, mark=none, dotted] table [x=x, y=mean, col sep=semicolon, row sep=newline, /pgf/number format/read comma as period]{\simarmadillorandomtranslation};
                    % Plot Active
                    \addplot [color=red, line width=1.5pt] table [x=x, y=mean, col sep=semicolon, row sep=newline, /pgf/number format/read comma as period]{\simarmadilloactivetranslation};
                    % Plot zero line
                    \draw[black, thin, densely dotted] (axis cs:\pgfkeysvalueof{/pgfplots/xmin},0) -- (axis cs:\pgfkeysvalueof{/pgfplots/xmax},0);
                    
                    %\node[anchor=north west, draw=black,fill=white] at (rel axis cs:0.02,0.98) {\footnotesize Armadillo};
                    
                % Plot Lucy
                \nextgroupplot
                    %\node [draw=black] at (rel axis cs:0.79, 0.81){\includegraphics[width=0.8cm, height=1cm]{figures/lucy_crop.png}};
                    % Plot Random 
                    \addplot [color=green!60!black, line width=1.5pt, mark=none, dotted] table [x=x, y=mean, col sep=semicolon, row sep=newline, /pgf/number format/read comma as period]{\simlucyrandomtranslation};
                    % Plot Active
                    \addplot [color=red, line width=1.5pt] table [x=x, y=mean, col sep=semicolon, row sep=newline, /pgf/number format/read comma as period]{\simlucyactivetranslation};
                    %\node[anchor=north west, draw=black,fill=white] at (rel axis cs:0.02,0.98) {\footnotesize Lucy};
                    
            \end{groupplot}
            \end{tikzpicture}

     \end{subfigure}
     
    \caption{Simulation experiments on five meshes from the Stanford Scanning Repository}
    \label{fig:sim_random_active}
\end{figure}
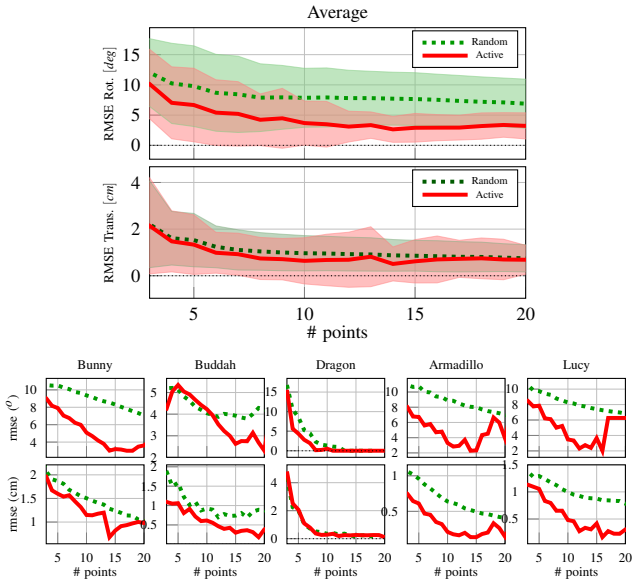

%% file: figures/real_sim_random_v_active.tex
%\pgfplotsset{every axis title/.append style={at={(0.1,0.8)}}}
\begin{figure}[t!]
    \centering
    \begin{subfigure}[b]{\columnwidth}
         \centering
             \begin{tikzpicture}
                \begin{groupplot}[group style={group size= 1 by 2, 
                x descriptions at=edge bottom, 
%                y descriptions at=edge left, 
                ylabels at=edge left, 
                xlabels at=edge bottom,
                horizontal sep=0.3cm,
                vertical sep=0.1cm
                },
                xmajorgrids=true,
                ymajorgrids=true,
                grid style={line width=.1pt, draw=gray!10},major grid style={line width=.2pt,draw=gray!50},
                xtick pos=bottom,
                ytick pos=left,
                xmin=3,
                xmax=20,
                ylabel shift = -5pt,
                xlabel shift = -5pt,
                height=0.25*\height,
                scale only axis,
                width=5cm,
                xlabel={\# touches},
                tick label style={font=\scriptsize},
                title style={yshift=-1.3ex, font=\scriptsize},
                legend style={nodes={scale=0.4, transform shape}},
                y label style={at={(0,0.5)}, font=\scriptsize, yshift=+2.0ex},
                label style={font=\scriptsize}
                ]
            
            % ================================LOADING=========================================
            %-- Load rotation
            %Random
            \pgfplotstableread[col sep=semicolon]{data/real_random_v_active/overall_rotation_random.csv}\realoverallrandomrotation;
            %Active
            \pgfplotstableread[col sep=semicolon]{data/real_random_v_active/overall_rotation_active.csv}\realoverallactiverotation;
            
            %-- Load translation
            %Random
            \pgfplotstableread[col sep=semicolon]{data/real_random_v_active/overall_translation_random.csv}\realoverallrandomtranslation;
            %Active
            \pgfplotstableread[col sep=semicolon]{data/real_random_v_active/overall_translation_active.csv}\realoverallactivetranslation;

            \nextgroupplot[ylabel=\tiny{RMSE Rot. $[deg]$}, title=Average, title style={text depth = 0pt}]
                \addplot [color=orange, line width=1.5pt, mark=none, dotted] table [x=x, y=mean, col sep=semicolon, row sep=newline, /pgf/number format/read comma as period]{\realoverallrandomrotation};
                \addplot [name path=r_down, color=orange!50, draw opacity=0.5, forget plot] table [x=x, y=lmean, col sep=semicolon, row sep=newline, /pgf/number format/read comma as period]{\realoverallrandomrotation};
                \addplot [name path=r_top, color=orange!50, draw opacity=0.5, forget plot] table [x=x, y=umean, col sep=semicolon, row sep=newline, /pgf/number format/read comma as period]{\realoverallrandomrotation};
                \addplot[orange!50,fill opacity=0.5, forget plot] fill between[of=r_top and r_down];
                
                \addplot [color=blue, line width=1.5pt, mark=none] table [x=x, y=mean, col sep=semicolon, row sep=newline, /pgf/number format/read comma as period]{\realoverallactiverotation};
                \addplot [name path=a_down, color=blue!50, draw opacity=0.5, forget plot] table [x=x, y=lmean, col sep=semicolon, row sep=newline, /pgf/number format/read comma as period]{\realoverallactiverotation};
                \addplot [name path=a_top, color=blue!50, draw opacity=0.5, forget plot] table [x=x, y=umean, col sep=semicolon, row sep=newline, /pgf/number format/read comma as period]{\realoverallactiverotation};
                \addplot[blue!50,fill opacity=0.5, forget plot] fill between[of=a_top and a_down];
                
                \draw[black, thin, densely dotted] (axis cs:\pgfkeysvalueof{/pgfplots/xmin},0) -- (axis cs:\pgfkeysvalueof{/pgfplots/xmax},0);
                \legend{Random, Active}
                % \addplot table [x=x, y=Active, col sep=semicolon, row sep=newline, /pgf/number format/read comma as period]{\mydata};
                % \legend{$Random$,$Active$}
                
            \nextgroupplot[ylabel=\tiny{RMSE Trans. $[cm]$}]
                \addplot [color=orange, line width=1.5pt, mark=none, dotted] table [x=x, y=mean, col sep=semicolon, row sep=newline, /pgf/number format/read comma as period]{\realoverallrandomtranslation};
                \addplot [name path=r_down, color=orange!50, draw opacity=0.5, forget plot] table [x=x, y=lmean, col sep=semicolon, row sep=newline, /pgf/number format/read comma as period]{\realoverallrandomtranslation};
                \addplot [name path=r_top, color=orange!50, draw opacity=0.5, forget plot] table [x=x, y=umean, col sep=semicolon, row sep=newline, /pgf/number format/read comma as period]{\realoverallrandomtranslation};
                \addplot[orange!50,fill opacity=0.5, forget plot] fill between[of=r_top and r_down];
                
                \addplot [color=blue, line width=1.5pt] table [x=x, y=mean, col sep=semicolon, row sep=newline, /pgf/number format/read comma as period]{\realoverallactivetranslation};
                \addplot [name path=a_down, color=blue!50, draw opacity=0.5, forget plot] table [x=x, y=lmean, col sep=semicolon, row sep=newline, /pgf/number format/read comma as period]{\realoverallactivetranslation};
                \addplot [name path=a_top, color=blue!50, draw opacity=0.5, forget plot] table [x=x, y=umean, col sep=semicolon, row sep=newline, /pgf/number format/read comma as period]{\realoverallactivetranslation};
                \addplot[blue!50,fill opacity=0.5, forget plot] fill between[of=a_top and a_down];
                
                \draw[black, thin, densely dotted] (axis cs:\pgfkeysvalueof{/pgfplots/xmin},0) -- (axis cs:\pgfkeysvalueof{/pgfplots/xmax},0);
                \legend{Random, Active}
                % \addplot table [x=x, y=Active, col sep=semicolon, row sep=newline, /pgf/number format/read comma as period]{\mydata};
                % \legend{$Random$,$Active$}
            \end{groupplot}
            \end{tikzpicture}

     \end{subfigure}
     \hfill
     \begin{subfigure}[b]{\columnwidth}
         \centering
         \begin{tikzpicture}
            \begin{groupplot}[group style={group size= 5 by 2,
                x descriptions at=edge bottom, 
%                y descriptions at=edge left, 
                ylabels at=edge left, 
                xlabels at=edge bottom,
                horizontal sep=0.3cm,
                vertical sep=0.1cm
                },
                xmajorgrids=true,
                ymajorgrids=true,
                grid style={line width=.1pt, draw=gray!10},major grid style={line width=.2pt,draw=gray!50},
                xtick pos=bottom,
                ytick pos=left,
                xmin=2,
                xmax=20,
                ylabel shift = -5pt,
                xlabel shift = -5pt,
                height=0.15*\height,
                scale only axis,
                width=1.3cm,
                xlabel={\# touches},
                tick label style={font=\tiny},
                title style={yshift=-1.3ex, font=\tiny},
                label style={font=\tiny}
                ]
                
                % ================================LOADING=========================================
                %-- Loading Cleaner
                % Rotation
                \pgfplotstableread[col sep=semicolon]{data/real_random_v_active/cleaner_random_rotation.csv}\realcleanerrandomrotation;
                \pgfplotstableread[col sep=semicolon]{data/real_random_v_active/cleaner_active_rotation.csv}\realcleaneractiverotation;
                % Translation
                \pgfplotstableread[col sep=semicolon]{data/real_random_v_active/cleaner_random_translation.csv}\realcleanerrandomtranslation;
                \pgfplotstableread[col sep=semicolon]{data/real_random_v_active/cleaner_active_translation.csv}\realcleaneractivetranslation;

                %-- Loading Olive Oil
                %Rotation
                \pgfplotstableread[col sep=semicolon]{data/real_random_v_active/oliveoil_random_rotation.csv}\realoliveoilrandomrotation;
                \pgfplotstableread[col sep=semicolon]{data/real_random_v_active/oliveoil_active_rotation.csv}\realoliveoilactiverotation;
                %Translation
                \pgfplotstableread[col sep=semicolon]{data/real_random_v_active/oliveoil_random_translation.csv}\realoliveoilrandomtranslation;
                \pgfplotstableread[col sep=semicolon]{data/real_random_v_active/oliveoil_active_translation.csv}\realoliveoilactivetranslation;

                %-- Loading shampoo
                %Rotation
                \pgfplotstableread[col sep=semicolon]{data/real_random_v_active/shampoo_random_rotation.csv}\realshampoorandomrotation;
                \pgfplotstableread[col sep=semicolon]{data/real_random_v_active/shampoo_active_rotation.csv}\realshampooactiverotation;
                %Translation
                \pgfplotstableread[col sep=semicolon]{data/real_random_v_active/shampoo_random_translation.csv}\realshampoorandomtranslation;
                \pgfplotstableread[col sep=semicolon]{data/real_random_v_active/shampoo_active_translation.csv}\realshampooactivetranslation;

                %-- Loading spray
                %Rotation
                \pgfplotstableread[col sep=semicolon]{data/real_random_v_active/spray_random_rotation.csv}\realsprayrandomrotation;
                \pgfplotstableread[col sep=semicolon]{data/real_random_v_active/spray_active_rotation.csv}\realsprayactiverotation;        
                %Translation
                \pgfplotstableread[col sep=semicolon]{data/real_random_v_active/spray_random_translation.csv}\realsprayrandomtranslation;
                \pgfplotstableread[col sep=semicolon]{data/real_random_v_active/spray_active_translation.csv}\realsprayactivetranslation;

                %-- Loading sugar
                %Rotation
                \pgfplotstableread[col sep=semicolon]{data/real_random_v_active/sugar_random_rotation.csv}\realsugarrandomrotation;
                \pgfplotstableread[col sep=semicolon]{data/real_random_v_active/sugar_active_rotation.csv}\realsugaractiverotation;
                %Translation
                \pgfplotstableread[col sep=semicolon]{data/real_random_v_active/sugar_random_translation.csv}\realsugarrandomtranslation;
                \pgfplotstableread[col sep=semicolon]{data/real_random_v_active/sugar_active_translation.csv}\realsugaractivetranslation;

                %horizontal sep=1mm,
                %vertical sep=1mm},
                % =========================================================================

                % =====================================Rotation=====================================
                % Plot Cleaner
                \nextgroupplot[ylabel=\tiny{rmse (${}^o$)}, y label style={at={(0,0.5)}, font=\tiny, yshift=+1.5ex}, title=Cleaner, title style={text depth = 0pt}]
                    %\node [draw=black] at (rel axis cs:0.8, 0.85){\includegraphics[width=0.8cm, height=0.8cm]{figures/bunny_crop.png}};
                    %\node [draw=black] at (rel axis cs:0.79, 0.81){\includegraphics[width=0.8cm, height=1cm]{figures/bunny_crop.png}};
                    % Plot Random
                    \addplot [color=orange, line width=1.5pt, mark=none, dotted] table [x=x, y=mean, col sep=semicolon, row sep=newline, /pgf/number format/read comma as period]{\realcleanerrandomrotation};
                    % Plot Active
                    \addplot [color=blue, line width=1.5pt] table [x=x, y=mean, col sep=semicolon, row sep=newline, /pgf/number format/read comma as period]{\realcleaneractiverotation};
                    
                    % Plot zero line
                    \draw[black, thin, densely dotted] (axis cs:\pgfkeysvalueof{/pgfplots/xmin},0) -- (axis cs:\pgfkeysvalueof{/pgfplots/xmax},0);
                    
                    %\node[anchor=north west, draw=black,fill=white] at (rel axis cs:0.02,0.98) {\footnotesize Bunny};
                    
                % Plot Olive Oil
                \nextgroupplot[title=Olive Oil, title style={text depth = 0pt}]
                    % Plot Random 
                    %\node [draw=black] at (rel axis cs:0.79, 0.81){\includegraphics[width=0.8cm, height=1cm]{figures/buddah_crop.png}};
                    \addplot [color=orange, line width=1.5pt, mark=none, dotted] table [x=x, y=mean, col sep=semicolon, row sep=newline, /pgf/number format/read comma as period]{\realoliveoilrandomrotation};
                    % Plot Active
                    \addplot [color=blue, line width=1.5pt] table [x=x, y=mean, col sep=semicolon, row sep=newline, /pgf/number format/read comma as period]{\realoliveoilactiverotation};
                    % Plot zero line
                    \draw[black, thin, densely dotted] (axis cs:\pgfkeysvalueof{/pgfplots/xmin},0) -- (axis cs:\pgfkeysvalueof{/pgfplots/xmax},0);
                    
                    %\node[anchor=north west, draw=black,fill=white] at (rel axis cs:0.02,0.98) {\footnotesize Buddha};
                    
                % Plot Shampoo
                \nextgroupplot[title=Shampoo, title style={text depth = 0pt}]
                    %\node [draw=black] at (rel axis cs:0.78, 0.85){\includegraphics[width=1cm, height=0.8cm]{figures/dragon_crop.png}};
                    %\node [draw=black] at (rel axis cs:0.79, 0.81){\includegraphics[width=0.8cm, height=1cm]{figures/dragon_crop.png}};
                    % Plot Random 
                    \addplot [color=orange, line width=1.5pt, mark=none, dotted] table [x=x, y=mean, col sep=semicolon, row sep=newline, /pgf/number format/read comma as period]{\realshampoorandomrotation};
                    % Plot Active
                    \addplot [color=blue, line width=1.5pt] table [x=x, y=mean, col sep=semicolon, row sep=newline, /pgf/number format/read comma as period]{\realshampooactiverotation};
                    % Plot zero line
                    \draw[black, thin, densely dotted] (axis cs:\pgfkeysvalueof{/pgfplots/xmin},0) -- (axis cs:\pgfkeysvalueof{/pgfplots/xmax},0);
                    %\node[anchor=north west, draw=black,fill=white] at (rel axis cs:0.02,0.98) {\footnotesize Dragon};
                    
                % Plot Spray
                \nextgroupplot[title=Spray, title style={text depth = 0pt}]
                    %\node [draw=black] at (rel axis cs:0.79, 0.81){\includegraphics[width=0.8cm, height=1cm]{figures/armadillo_crop.png}};
                    % Plot Random 
                    \addplot [color=orange, line width=1.5pt, mark=none, dotted] table [x=x, y=mean, col sep=semicolon, row sep=newline, /pgf/number format/read comma as period]{\realsprayrandomrotation};
                    % Plot Active
                    \addplot [color=blue, line width=1.5pt] table [x=x, y=mean, col sep=semicolon, row sep=newline, /pgf/number format/read comma as period]{\realsprayactiverotation};
                    % Plot zero line
                    \draw[black, thin, densely dotted] (axis cs:\pgfkeysvalueof{/pgfplots/xmin},0) -- (axis cs:\pgfkeysvalueof{/pgfplots/xmax},0);
                    
                    %\node[anchor=north west, draw=black,fill=white] at (rel axis cs:0.02,0.98) {\footnotesize Armadillo};
                    
                % Plot Sugar
                \nextgroupplot[title=Sugar, title style={text depth = 0pt}]
                    %\node [draw=black] at (rel axis cs:0.79, 0.81){\includegraphics[width=0.8cm, height=1cm]{figures/lucy_crop.png}};
                    % Plot Random 
                    \addplot [color=orange, line width=1.5pt, mark=none, dotted] table [x=x, y=mean, col sep=semicolon, row sep=newline, /pgf/number format/read comma as period]{\realsugarrandomrotation};
                    % Plot Active
                    \addplot [color=blue, line width=1.5pt] table [x=x, y=mean, col sep=semicolon, row sep=newline, /pgf/number format/read comma as period]{\realsugaractiverotation};
                    %\node[anchor=north west, draw=black,fill=white] at (rel axis cs:0.02,0.98) {\footnotesize Lucy};

            % =====================================Translation=====================================
                % Plot Cleaner
                \nextgroupplot[ylabel=\tiny{rmse (cm)}]
                    %\node [draw=black] at (rel axis cs:0.8, 0.85){\includegraphics[width=0.8cm, height=0.8cm]{figures/bunny_crop.png}};
                    %\node [draw=black] at (rel axis cs:0.79, 0.81){\includegraphics[width=0.8cm, height=1cm]{figures/bunny_crop.png}};
                    % Plot Random
                    \addplot [color=orange, line width=1.5pt, mark=none, dotted] table [x=x, y=mean, col sep=semicolon, row sep=newline, /pgf/number format/read comma as period]{\realcleanerrandomtranslation};
                    % Plot Active
                    \addplot [color=blue, line width=1.5pt] table [x=x, y=mean, col sep=semicolon, row sep=newline, /pgf/number format/read comma as period]{\realcleaneractivetranslation};
                    
                    % Plot zero line
                    \draw[black, thin, densely dotted] (axis cs:\pgfkeysvalueof{/pgfplots/xmin},0) -- (axis cs:\pgfkeysvalueof{/pgfplots/xmax},0);
                    
                    %\node[anchor=north west, draw=black,fill=white] at (rel axis cs:0.02,0.98) {\footnotesize Bunny};
                    
                % Plot Olive Oil 
                \nextgroupplot
                    % Plot Random 
                    %\node [draw=black] at (rel axis cs:0.79, 0.81){\includegraphics[width=0.8cm, height=1cm]{figures/buddah_crop.png}};
                    \addplot [color=orange, line width=1.5pt, mark=none, dotted] table [x=x, y=mean, col sep=semicolon, row sep=newline, /pgf/number format/read comma as period]{\realoliveoilrandomtranslation};
                    % Plot Active
                    \addplot [color=blue, line width=1.5pt] table [x=x, y=mean, col sep=semicolon, row sep=newline, /pgf/number format/read comma as period]{\realoliveoilactivetranslation};
                    % Plot zero line
                    \draw[black, thin, densely dotted] (axis cs:\pgfkeysvalueof{/pgfplots/xmin},0) -- (axis cs:\pgfkeysvalueof{/pgfplots/xmax},0);
                    
                    %\node[anchor=north west, draw=black,fill=white] at (rel axis cs:0.02,0.98) {\footnotesize Buddha};
                    
                % Plot Shampoo
                \nextgroupplot
                    %\node [draw=black] at (rel axis cs:0.78, 0.85){\includegraphics[width=1cm, height=0.8cm]{figures/dragon_crop.png}};
                    %\node [draw=black] at (rel axis cs:0.79, 0.81){\includegraphics[width=0.8cm, height=1cm]{figures/dragon_crop.png}};
                    % Plot Random 
                    \addplot [color=orange, line width=1.5pt, mark=none, dotted] table [x=x, y=mean, col sep=semicolon, row sep=newline, /pgf/number format/read comma as period]{\realshampoorandomtranslation};
                    % Plot Active
                    \addplot [color=blue, line width=1.5pt] table [x=x, y=mean, col sep=semicolon, row sep=newline, /pgf/number format/read comma as period]{\realshampooactivetranslation};
                    % Plot zero line
                    \draw[black, thin, densely dotted] (axis cs:\pgfkeysvalueof{/pgfplots/xmin},0) -- (axis cs:\pgfkeysvalueof{/pgfplots/xmax},0);
                    %\node[anchor=north west, draw=black,fill=white] at (rel axis cs:0.02,0.98) {\footnotesize Dragon};
                    
                % Plot Spray
                \nextgroupplot
                    %\node [draw=black] at (rel axis cs:0.79, 0.81){\includegraphics[width=0.8cm, height=1cm]{figures/armadillo_crop.png}};
                    % Plot Random 
                    \addplot [color=orange, line width=1.5pt, mark=none, dotted] table [x=x, y=mean, col sep=semicolon, row sep=newline, /pgf/number format/read comma as period]{\realsprayrandomtranslation};
                    % Plot Active
                    \addplot [color=blue, line width=1.5pt] table [x=x, y=mean, col sep=semicolon, row sep=newline, /pgf/number format/read comma as period]{\realsprayactivetranslation};
                    % Plot zero line
                    \draw[black, thin, densely dotted] (axis cs:\pgfkeysvalueof{/pgfplots/xmin},0) -- (axis cs:\pgfkeysvalueof{/pgfplots/xmax},0);
                    
                    %\node[anchor=north west, draw=black,fill=white] at (rel axis cs:0.02,0.98) {\footnotesize Armadillo};
                    
                % Plot Sugar
                \nextgroupplot
                    %\node [draw=black] at (rel axis cs:0.79, 0.81){\includegraphics[width=0.8cm, height=1cm]{figures/lucy_crop.png}};
                    % Plot Random 
                    \addplot [color=orange, line width=1.5pt, mark=none, dotted] table [x=x, y=mean, col sep=semicolon, row sep=newline, /pgf/number format/read comma as period]{\realsugarrandomtranslation};
                    % Plot Active
                    \addplot [color=blue, line width=1.5pt] table [x=x, y=mean, col sep=semicolon, row sep=newline, /pgf/number format/read comma as period]{\realsugaractivetranslation};
                    %\node[anchor=north west, draw=black,fill=white] at (rel axis cs:0.02,0.98) {\footnotesize Lucy};
                    
            \end{groupplot}
            \end{tikzpicture}

     \end{subfigure}
     
    \caption{Robot experiments on five selected daily objects}
    \label{fig:real_random_active}
\end{figure}
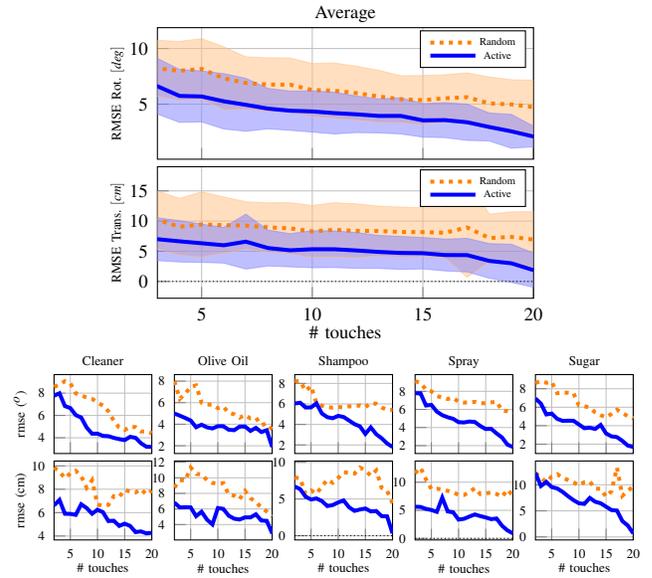
%% mka efigure smal;ler

%% file: sections/conclusions.tex
\section{CONCLUSIONS}
\label{sec:conclusions}
In this paper we proposed a novel active visuo-tactile framework for object pose estimation. The robotic system using our proposed active touch-based approach guided by a vision estimate, accurately and efficiently estimates the pose of objects in an unknown workspace. Moreover, the vision estimate is corrected by using the tactile modality. Furthermore, our proposed method enables the robotic system to \textit{actively} reasons upon possible next actions and choose the next best touch based on an information gain metric. We compared the performance of our framework with random touch point acquisition and active touch point acquisition. We demonstrated that using the active touch point selection, on average highly accurate results can be achieved with fewer measurements. We validated our method in simulation and a robotic system. \newline As future work, we would like to evaluate our method for the time-variant $SE(3)$ estimation wherein the objects can even move during exploration. We will also relax the assumption for the need of an accurate model mesh for the objects of interest.
Furthermore, we will extend our visuo-tactile approach to deformable objects with dynamic center of mass thus relaxing the assumption of rigid objects~\cite{kaboli2016tactilehuman, yao2017tactile}.
Finally, we will enable our robotic system with a complex motion planning algorithm to explore the unknown workspace autonomously via vision and tactile sensing. 

%% file: sections/acknowledgement.tex
We would like to sincerely thank Dr. V\"ogel, Dr. Enders, and Mr. Anzuela  for assistance in the experimental setup. 